\documentclass[10pt,twocolumn,letterpaper]{article}

\usepackage{cvpr}              

\usepackage{graphicx}
\usepackage{amsmath}
\usepackage{amssymb}
\usepackage{booktabs}
\usepackage{tabularx}
\usepackage{xcolor}
\usepackage{colortbl}

\usepackage[pagebackref,breaklinks,colorlinks]{hyperref}

\usepackage[capitalize]{cleveref}
\crefname{section}{Sec.}{Secs.}
\Crefname{section}{Section}{Sections}
\Crefname{table}{Table}{Tables}
\crefname{table}{Tab.}{Tabs.}

\def\cvprPaperID{4606} 
\def\confName{CVPR}
\def\confYear{2023}

\graphicspath{{pics}{../pics}}
\DeclareGraphicsExtensions{.pdf,.jpeg,.png}

\usepackage{makecell,multirow}
\usepackage{algorithmic}
\usepackage[ruled,vlined]{algorithm2e}
\usepackage{epsfig}
\usepackage{enumitem}
\usepackage{color}
\usepackage{float} 
\usepackage{ragged2e}
\usepackage[english]{babel}
\usepackage{tabularx}
\usepackage{subcaption}

\newcommand{\nothing}[1]{}
\definecolor{DeltaColor}{rgb}{0.039,0.73,0.71}
\definecolor{SetaColor}{rgb}{0.867, 0.0235, 0.376}
\definecolor{SigmaColor}{rgb}{0.98,0.45,0.0}
\definecolor{RedColor}{rgb}{0.8,0,0}
\definecolor{AlphaColor}{rgb}{0,0,0.8}
\definecolor{BetaColor}{rgb}{0.8,0,0.8}
\definecolor{GammaColor}{rgb}{0.5,0,0.7}
\definecolor{EpsilonColor}{rgb}{0.353,0.725,0.906}
\definecolor{TauColor}{rgb}{0.423,0.235,0.192}
\definecolor{WtColor}{rgb}{0.235,0.470,0.470}
\newcommand{\red}[1]{\textbf{\color{red} #1}}
\newcommand{\green}[1]{\textbf{\color{green} #1}}
\newcommand{\blue}[1]{\textbf{\color{blue} #1}}
\newcommand{\bcr}[1]{\textbf{\color{red} #1}}
\newcommand{\bcg}[1]{\textbf{\color{green} #1}}
\newcommand{\bcb}[1]{\textbf{\color{blue} #1}}

\definecolor{AudioColor}{rgb}{0.56,0.34,0.62}

\definecolor{DeadlineColor}{rgb}{0.9,0.4,0} 

\definecolor{figred}{rgb}{1,0,0}
\definecolor{figgreen}{rgb}{0,0.6,0}
\definecolor{figblue}{rgb}{0,0,1}
\definecolor{figpink}{rgb}{1,0.63,0.63}

\newcounter{pccount}
\newcommand{\filename}[1]{\url{#1}}
\newcommand{\foldername}[1]{\url{#1}}

\begin{document}

\title{
    Revisiting Long-tailed Image Classification: \\ Survey and Benchmarks with New Evaluation Metrics
    }
\author{
    Chaowei Fang$^1$ \;
    Dingwen Zhang$^2$ \;
    Wen Zheng$^3$ \;
    Xue Li$^1$ \;
    Le Yang$^2$  \;
    Lechao Cheng$^4$ \; 
    Junwei Han$^2$
    \\
    $^1$Xidian University\quad\quad
    $^2$Northwestern Polytechnical University\\
    $^3$China University of Mining Technology-Beijing\quad\quad
    $^4$Zhejiang Lab
}
\maketitle

\begin{abstract}
   Recently, long-tailed image classification harvests lots of research attention, since the data distribution is long-tailed in many real-world situations. Piles of algorithms are devised to address the data imbalance problem by biasing the training process towards less frequent classes. However, they usually evaluate the performance on a balanced testing set or multiple independent testing sets having distinct distributions with the training data. Considering the testing data may have arbitrary distributions, existing evaluation strategies are unable to reflect the actual classification performance objectively. We set up novel evaluation benchmarks based on a series of testing sets with evolving distributions. A corpus of metrics are designed for measuring the accuracy, robustness, and bounds of algorithms for learning with long-tailed distribution. Based on our benchmarks,  we re-evaluate the performance of existing methods on CIFAR10 and CIFAR100 datasets, which is valuable for guiding the selection of data rebalancing techniques. 
   We also revisit existing methods and categorize them into four types including data balancing, feature balancing, loss balancing, and prediction balancing, according the focused procedure during the training pipeline. 

\end{abstract}

\section{Introduction}

Due to the rapid development of deep convolutional neural networks (CNN)~\cite{simonyan2014very,he2016deep}, substantial progress is achieved in image recognition. 
However, they usually assume that the training and testing data is balanced. 
In practice, training or testing data appears to be long-tailed, e.g., there exist few samples for rare diseases in medical diagnosis~\cite{zhang2020exploring,zhang2021cross,zhao2021deep,zhao2021contralaterally,pan2022computer,zhang2021automatic} or endangered animals in species classification~\cite{zhang2022generalized,zhang2022onfocus,cheng2022compound}. 
As mentioned by~\cite{zhang2021weakly}, the case becomes even worse in weakly and semi-supervised learning scenarios~\cite{zhang2018spftn,zhang2019leveraging,zhang2020weakly,zhao2021weakly,huang2021scribble,pan2022learning,zhao2022cross,wang2022double,zhang2022generalized}.
The conventional training process of CNNs is dominated by frequent classes while rare classes are neglected. 
A large number of methods focus on rebalancing the training data through biasing the training process towards rare classes~\cite{cao2019learning,hong2021disentangling}.
However, they usually assume the testing data is balanced, while the distribution of real testing data is unknown and arbitrary. The performance of existing algorithms for learning with long-tailed distribution (LLTD) remains to be validated under such circumstance. 
This paper revisits existing evaluation strategies for long-tailed image classification and provides a new strategy to evaluate LLTD algorithms on testing data with unknown distribution.

\label{sec:intro}
\begin{figure}[t]
    \centering
    \includegraphics[width=1.0\linewidth]{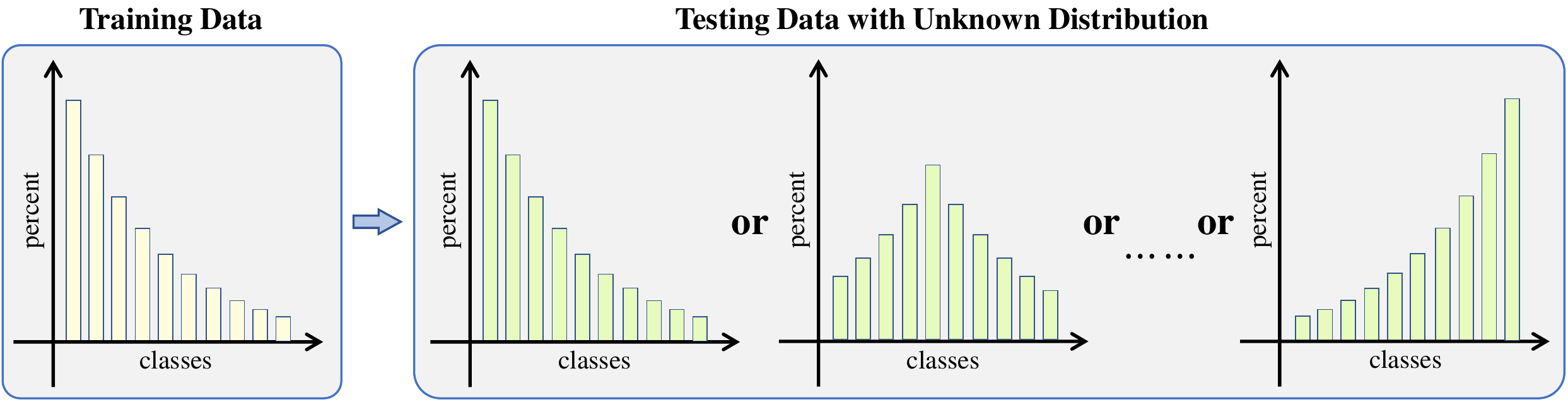}
    \caption{In real-world tasks for learning with long-tailed distribution, the distribution of testing data is unknown and may be different from that of training data.} 
    \label{fig:motivation}
\end{figure}

\begin{table*}[t]
    \fontsize{7.5}{9} \selectfont
    \centering
    \caption{Evaluation strategies for long-tailed image classification, including: 1) testing algorithms on balanced subset~\cite{zhou2020bbn,park2021influence}; 2) testing algorithms on subsets having uniform distribution and forward/backward-trend distribution~\cite{zhang2021test}; 3) testing algorithms on a series of subsets with dynamic evolving distributions. }
    \setlength{\tabcolsep}{1mm}{
    \begin{tabular}{r|l|l|r}
    \toprule
    \toprule
    \multicolumn{1}{l|}{\textbf{Test Data}} & \multicolumn{1}{l}{\textbf{Metrics}} & \textbf{Descriptions} & \multicolumn{1}{l}{\textbf{Evaluation Properties}} \\
    \midrule
    \midrule
          & \cellcolor[rgb]{ .949,  .949,  .949}ACC$_{all}$ & \cellcolor[rgb]{ .949,  .949,  .949}Accuracy of all classes & \multicolumn{1}{l}{\cellcolor[rgb]{ .949,  .949,  .949}1. Only reflect the classification performance under a} \\
    \multicolumn{1}{l|}{Balanced} & ACC$_{many}$ & Accuracy of many-shot classes & \multicolumn{1}{l}{specific testing distribution.} \\
    \multicolumn{1}{l|}{Distribution} & \cellcolor[rgb]{ .949,  .949,  .949}ACC$_{mid}$ & \cellcolor[rgb]{ .949,  .949,  .949}Accuracy of medium-shot classes & \multicolumn{1}{l}{\cellcolor[rgb]{ .949,  .949,  .949}2. Not able to reflect the stability and upper/lower} \\
          & ACC$_{few}$ & Accuracy of few-shot classes & \multicolumn{1}{l}{bound of algorithms on real-world testing data.} \\
    \midrule
    \midrule
    \multicolumn{1}{l|}{Multiple} & \cellcolor[rgb]{ .949,  .949,  .949}ACC$_{forw}$ & \cellcolor[rgb]{ .949,  .949,  .949}Accuracy under forward-trend test distribution & \multicolumn{1}{l}{\cellcolor[rgb]{ .949,  .949,  .949}1. Reflect the classification performance under several} \\
    \multicolumn{1}{l|}{Independent} & ACC$_{uni}$ & Accuracy under uniform test distribution & \multicolumn{1}{l}{dependent imbalanced testing distributions.} \\
    \multicolumn{1}{l|}{Distributions} & \cellcolor[rgb]{ .949,  .949,  .949}ACC$_{back}$ & \cellcolor[rgb]{ .949,  .949,  .949}Accuracy under backward-trend test distribution & \multicolumn{1}{l}{\cellcolor[rgb]{ .949,  .949,  .949}2. Coarsely reflect stability and upper/lower bound.} \\
    \midrule
    \midrule
          & AUC   & Area under accuracy curve &  \\
    \multicolumn{1}{l|}{Dynamic} & \cellcolor[rgb]{ .949,  .949,  .949}ACC$_{avg}$ & \cellcolor[rgb]{ .949,  .949,  .949}Average accuracy & \multicolumn{1}{l}{\cellcolor[rgb]{ .949,  .949,  .949}1. Reflect the classification performance under real} \\
    \multicolumn{1}{l|}{Evolving} & ACC$_{std}$ & Standard deviation of accuracy & \multicolumn{1}{l}{testing data more comprehensively.} \\
    \multicolumn{1}{l|}{Distributions} & \cellcolor[rgb]{ .949,  .949,  .949}ACC$_{max}$ & \cellcolor[rgb]{ .949,  .949,  .949}Maximum accuracy & \multicolumn{1}{l}{\cellcolor[rgb]{ .949,  .949,  .949}2. Evaluate stability and upper/lower bound more finely.} \\
          & ACC$_{min}$ & Minimum accuracy &  \\
          & DR    & Drop ratio of accuracy &  \\
    \bottomrule
    \bottomrule
    \end{tabular}%
    }
    
    \label{tab:comp-eval}
\end{table*}

Recently, learning models with long-tailed training data attracts lots of research interest. 
Existing methods concentrate on different procedures including data preparation, feature representation learning, objective function design, and class prediction to tackle this task.
According to the focused procedures, they can be categorized into four types, i.e., data balancing~\cite{chawla2002smote,drummond2003c4}, feature balancing~\cite{cui2021parametric}, loss balancing~\cite{lin2017focal,park2021influence,ren2020balanced}, and prediction balancing~\cite{wang2020long,zhang2021test,li2022trustworthy}.
Most of them train models with imbalanced subsets of existing datasets, such as CIFAR-10/100~\cite{krizhevsky2009learning}, ImageNetLT~\cite{liu2019large}, and iNaturalist~\cite{van2018inaturalist}. Then, they evaluate the classification performance on a balanced testing set with accuracy values of all classes and partial classes (e.g., many/middle/few-shot classes). 
This evaluation process differs from the real-world situation where the testing data distribution is unknown. 
Hence, it may not reflect the actual classification performance objectively.
Moreover, those metrics can not indicate the stability and performance bounds of algorithms on testing data which may have arbitrary distributions.
\cite{zhang2021test} attempts to estimate LLTD algorithms on testing data with multiple distributions: 1) Testing data has an imbalanced distribution sharing the forward trend as the training data; 2) Testing data has uniform sample sizes across classes, namely the distribution is balanced; 3) Testing data has imbalanced distributions with the backward trend of the training data. However, this manner is still limited in comprehensively evaluating LLTD algorithms on unknown testing data. It can only coarsely reflect algorithmic stability and bounds.

To address the above issues, we devise new evaluation metrics based on testing data with dynamic evolving distributions. For covering possible test distributions as fully as possible, we simulate a series of testing sets by shifting the frequent classes according to the class index. Then, each case of testing data is used to calculate the accuracy of LLTD algorithms. A corpus of evaluation metrics are estimated from the accuracy on all testing sets: 1) The area under the curve formed by those accuracy values and the average accuracy can be used for evaluating the performance on universal testing data; 2) The standard deviation and the accuracy drop ratio reflect the stability against the variation of test data distributions; 3) The maximum and minimum accuracy values can approximate the upper and lower bound, respectively. The comparison between our evaluation strategy and existing strategies is provided in Table~\ref{tab:comp-eval}.

Based on the new evaluation metrics, we provide an elaborated analysis about the classification accuracy, robustness, and bounds for existing LLTD algorithms on two benchmarks in which the testing data has same or different imbalance ratio with the training data.
This can guide the selection of data rebalancing techniques.
We observe that a few methods such as the contrastive representation learning algorithm in~\cite{cui2021parametric},  ensembling models learned with different distributions~\cite{zhang2021test}, logit rebalancing strategies~\cite{ren2020balanced,li2022long}, and data distribution disentangling~\cite{hong2021disentangling} have distinguished performance on addressing the long-tailed learning problem.
However, there still exists large space for improving the classification of tail classes under large distribution shift or large-scale number of classes. Meanwhile, middle classes also need attentions during training.

Main contributions of this paper are as follows.

\begin{itemize}
\item[1)] We provide a simple survey for existing long-tailed learning algorithms and classify them into four types according to the key procedures during the pipeline for tackling the long-tailed learning, i.e., data rebalancing, loss design, feature representation learning, and category prediction.
\item[2)] We design a corpus of new evaluation metrics for analyzing the classification accuracy, stability, and bounds of LLTD algorithms on testing data with dynamic evolving distributions more comprehensively.
\item[3)] Based on the new evaluation metrics, two benchmarks where training data and testing data have same or different imbalance ratios are set up to evaluate existing LLTD algorithms.
\end{itemize}

\begin{figure*}[thbp]
    \centering
    \includegraphics[width=1\linewidth]{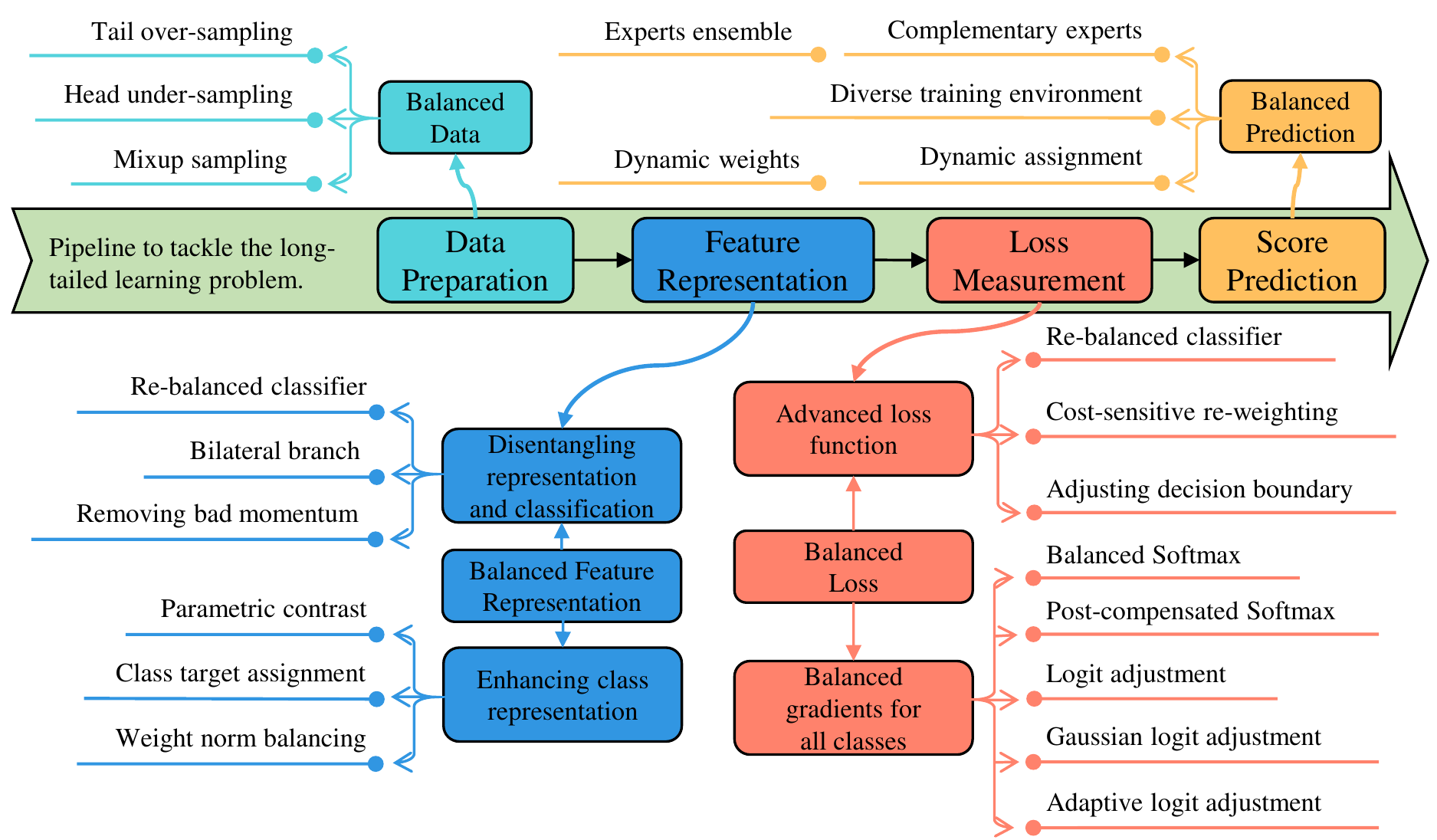}
    \put(-482,278){\cite{chawla2002smote}}
    \put(-491,256){\cite{drummond2003c4}}
    \put(-479,233){\cite{zhang2017mixup}}
    \put(-310,278){\cite{wang2020long}}
    \put(-232,256){\cite{tang2022invariant}}
    \put(-307,233){\cite{zhang2021test}}
    \put(-207,278){\cite{cai2021ace}}
    \put(-200,233){\cite{li2022trustworthy}}
    \put(-473,156){\cite{kang2019decoupling}}
    \put(-452,133){\cite{zhou2020bbn}}
    \put(-492,111){\cite{tang2020long}}
    \put(-457,78){\cite{cui2021parametric}}
    \put(-480,55){\cite{li2022targeted}}
    \put(-475,32){\cite{alshammari2022long}}
    \put(-51,172){\cite{lin2017focal}}
    \put(-34,146){\cite{park2021influence}}
    \put(-30,119){\cite{cao2019learning}}
    \put(-68,97){\cite{ren2020balanced}}
    \put(-34,77){\cite{hong2021disentangling}}
    \put(-72,55){\cite{menon2020long}}
    \put(-38,34){\cite{li2022long}}
    \put(-38,11){\cite{zhao2022adaptive}}
    \vspace{-5pt}
    \caption{Categorization of existing long-tailed image classification works. Based on four primary procedures, existing works can be categorized into four groups: balanced data, balanced feature representation, balanced loss, and balanced prediction.}
    \label{figCate}
\end{figure*}
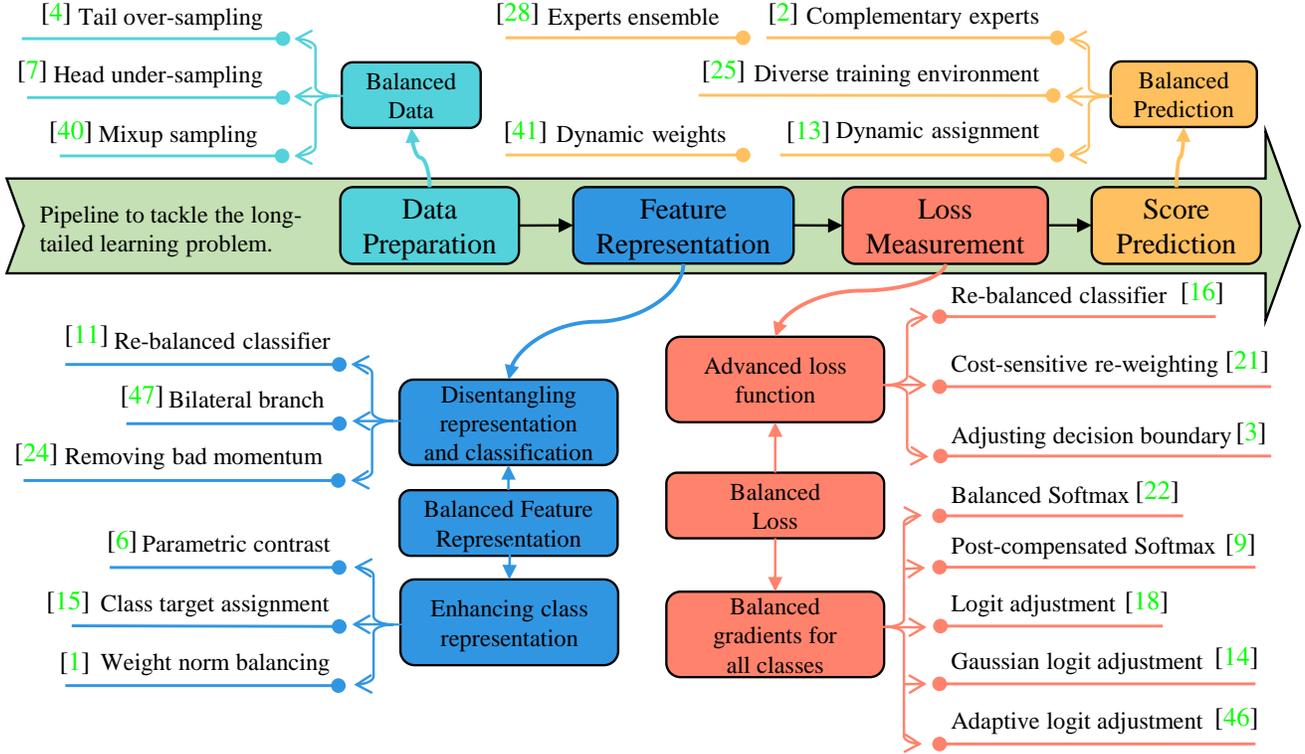

\section{Survey of Long-tailed Learning}
\label{sec:related}
\subsection{Problem Definition}
\label{sec:prob_def}
Real-world data often has a long-tailed distribution.
Suppose the training data be $\mathbb D^{trn}=\{(x_n, y_n)\}_{n=1}^{N^{trn}}$. $x_n$ and $y_n$ denote the $n$-th training image and its class label, respectively; $N^{trn}$ represents the number of training samples. 
Let the number of training samples belonging the $c$-th class be $N^{trn}_c$. The imbalance ratio of the training data is denoted as $\rho^{trn}=\max_c N^{trn}_c / \min_c N^{trn}_c$.
Suppose there exist $C$ target classes, i.e., $y_n\in[1,C]$. 
We define the testing dataset as  $\mathbb D^{tst}=\{(x_n, y_n)\}_{n=1}^{N^{tst}}$.
$N^{tst}$ denotes the number of testing samples. The number of testing samples belonging the $c$-th class is $N^{tst}_c$. The imbalance ratio of the testing data is denoted as $\rho^{tst}=\max_c N^{tst}_c/\min_c N^{tst}_c$.

We define the distribution shift between training and testing data with the JS divergence, 
\begin{equation}
\delta(\mathbb D^{trn}, \mathbb D^{tst})=-\sum_{c=1}^C [q_c^{trn} \ln(\frac{q_c^{trn}}{q_c^{tst}})+q_c^{tst} \ln(\frac{q_c^{tst}}{q_c^{trn}})],
\end{equation}
where $q_c^{trn}=\frac{N_c^{trn}}{N^{trn}}$ and $q_c^{tst}=\frac{N_c^{tst}}{N^{tst}}$.

Given the training data $\mathbb D^{trn}$, the goal is to learning a CNN model which can well adapt to the testing data $\mathbb D^{tst}$. We define the inference process of the CNN model as $f(\cdot)$. Namely, given an image $x$, the CNN model can generate a classification probability vector $\mathbf p\in \mathbb R^C$. The inferred label is denoted as $\hat{y}=\arg\max_c p[c]$. Here, $p[c]$ indicates the $c$-th class's probability value.

\subsection{A Survey of Prior Works}
Under long-tailed distribution, head classes are prone to dominate the learning process, thus impairing the accuracy of tail classes. The core factor to address this problem is balancing the learning process. As shown in Figure~\ref{figCate}, we categorize existing works into four groups: balanced data, balanced feature representation, balanced loss, and balanced prediction.

\subsubsection{Balanced Data}
A straightforward method to address the data imbalance problem is to construct balanced data distribution, i.e., re-sampling the training data. However, excessively sampling tail classes~\cite{chawla2002smote} induces the over-fitting issue, while under-sampling head classes~\cite{drummond2003c4} hampers the representation learning and weakens the accuracy of head classes. Based on the effectiveness of mixup-based methods \cite{zhang2017mixup, yun2019cutmix}, combining samples from different classes can alleviate the long-tailed challenge. However, a naive implementation is prone to generate more head-head pairs. To this end, Xu \etal \cite{Xu2021TowardsCM} propose a balance-oriented mixup algorithm by biasing the mixing factor towards tail classes and increasing the occurrence of head-tail pairs.

\subsubsection{Balanced Feature Representation}
Designing algorithms to learn balanced feature representations is the other promising direction for addressing the long-tailed learning problem. cRT~\cite{kang2019decoupling} finds that data imbalancement does not impair the representation ability. Hence, cRT~\cite{kang2019decoupling} learns the feature extraction backbone with the conventional training strategy and then employ the data re-balancing algorithms to train the classifier. Similarly, BBN~\cite{zhou2020bbn} unifies the traditional and re-balanced data sampling strategies, and gradually shifts focus from the former to the latter. Aiming to jointly acquire representative features and discriminative classification scores, Tang \etal \cite{tang2020long} use the moving average of momentum to measure the misleading effect of head classes during training, and build a causal inference algorithm to remove the misleading effect during inference.

Another feature balancing approach is to enhance the representation ability for each class, \eg by contrastive learning. 
Cui \etal \cite{cui2021parametric} propose to explicitly learn a feature center for each class, which is used to increase the inter-class separability. 
Li \etal \cite{li2022targeted} reveal that the imbalanced sample distribution leads to close feature centers for tail classes. Thus, they propose to constrain feature centers to be uniformly distributed. 
Alshammari \etal \cite{alshammari2022long} tackle the long-tailed challenge from the perspective of weight balancing, and apply the weight decay strategy to penalize large weights.

\subsubsection{Balanced Loss}
Another reasonable approach to the imbalance challenge is assigning relatively higher attention to tail classes during the network optimization process. For example, Focal loss~\cite{lin2017focal} assigns larger weights to samples with lower prediction confidences, \ie, the so-called hard samples. However, if the dataset exhibits severe imbalance, excessively emphasizing tail classes would lead to an over-fitting dilemma. To alleviate this challenge, Park \etal \cite{park2021influence} propose to re-weight samples according to the reverse of their influences on decision boundaries. LDAM~\cite{cao2019learning} tackles the data imbalance challenge by increasing margins of tail classes' decision boundaries, considering decision boundaries of head classes are more reliable than those of tail classes. 

The other line of works focus on balancing gradients for head and tail classes by adjusting the \textit{Softmax} function. Ren \etal \cite{ren2020balanced} make an early attempt to balance the \textit{Softmax} function and develop the meta-sampling strategy to dynamically adjust the data distribution in the training process. LADE~\cite{hong2021disentangling} proposes the post-compensated \textit{Softmax} strategy to disentangle the source data distribution from network predictions. In addition, Menon \etal~\cite{menon2020long} introduce the logit adjustment strategy. If testing samples obey the independent and identical distribution of training samples, the logit adjustment strategy can generate accurate predictions. However, it is impractical to guarantee the independent and identical distribution between training and testing samples. Moreover, the underlying distribution of testing samples is usually unknown. To cope with this problem, GCL~\cite{li2022long} introduces Gaussian clouds into the logit adjustment process, and adaptively sets the cloud size according to the sample size of each class. Recently, Zhao \etal~\cite{zhao2022adaptive} reveal that previous logit adjustment techniques primarily focus on the sample quantity of each class, while ignoring the difficulty of samples. Thus, Zhao \etal \cite{zhao2022adaptive} propose to prevent the over-optimization on easy  samples of tail classes, while highlighting the training on difficult samples of head classes.

\subsubsection{Balanced Prediction}
Another type of methods try to tackle the long-tailed challenge by improving the inference process. 
Aiming to balance the accuracy of head and tail classes, Wang \etal~\cite{wang2020long} learn multiple experts simultaneously and ensemble their predictions to reduce the bias of single models. A dynamic routing module is developed to control the computational costs. 
ACE~\cite{cai2021ace} attempts to learn multiple complementary expert models. Specifically, each expert model is responsible for distinguishing a specific set of classes while its responses to non-assigned classes are suppressed. 
Considering the real-world testing data may exhibit a distinct distribution compared to the training data, Zhang \etal \cite{zhang2021test} learn multiple models under different distributions and combines them with weights generated via testing-time adaptation. 
For decreasing the computational cost, Li \etal~\cite{li2022trustworthy} propose to measure the uncertainty of each expert, and assign experts to each sample dynamically. 
Tang \etal~\cite{tang2022invariant} utilize uniform intra-class data sampling and confidence-aware data sampling strategies to construct different training environments for learning features invariant to diversified attributes.

\section{The New Evaluation Metrics}
\label{sec:data}

\begin{table*}[t]
      \caption{Peformance of existing methods on CIFAR10 and CIFAR100 with $\rho^{trn}=\rho^{tst}=0.01$. $\uparrow$ ($\downarrow$) means the larger (smaller) metric value indicates better performance. \red{Red}, \green{green}, and \blue{blue} digits indicate the best, the second best, and the third best method, respectively.  `AUC': area under accuracy curve; `ACC': accuracy; `AVG': average; `STD': standard deviation; `MAX': maximum; `MIN': minimum; `BTD': balanced testing distribution.}\vspace{-5pt}
      \label{tab:CIFAR-01}
      \fontsize{7.5}{9} \selectfont
      \centering
      \setlength{\tabcolsep}{1.5mm}{
      \begin{tabular}{l|c|c|c|c|c|c|c|c|c|c|c|c|c|c}
        \hline
        \multirow{3}{*}{\textbf{Method}} & \multicolumn{7}{c|}{\bf CIFAR10} & \multicolumn{7}{c}{\bf CIFAR100} \\ \cline{2-15}
        & \multirow{2}{*}{\textbf{AUC}$\uparrow$} & \multicolumn{6}{c|}{\textbf{ACC}} & \multirow{2}{*}{\textbf{AUC}$\uparrow$} & \multicolumn{6}{c}{\textbf{ACC}} \\ \cline{3-8} \cline{10-15}
        & & AVG$\uparrow$ & STD$\downarrow$ & MAX$\uparrow$ & MIN$\uparrow$ & DR$\downarrow$ & BTD$\uparrow$ & & AVG$\uparrow$ & STD$\downarrow$ & MAX$\uparrow$ & MIN$\uparrow$ & DR$\downarrow$ & BTD$\uparrow$ \\ 
        \hline
        CE & 
        68.07 & 67.57 & 14.55 & 89.32 & 45.69 & 48.8\% & 67.67 &
        36.48 & 37.25 & 16.31 & 60.71 & 14.38 & 76.3\% & 37.26 \\
        Focal~\cite{lin2017focal}  & 
        71.19 & 70.89 & 12.63 & \blue{89.98} & 52.98 & 41.1\% & 70.40 &
        36.99 & 37.75 & 17.34 & 62.47 & 13.52 & 78.4\% & 37.65 \\
        LDAM~\cite{cao2019learning} & 
        75.35 & 75.42 &  8.39 & 89.75 & 63.19 & 29.6\% & 75.26 &
        42.51 & 42.80 & 12.33 & 59.98 & 25.54 & 57.4\% & 42.77 \\
        cRT~\cite{kang2019decoupling} & 
        68.08 & 68.05 & 12.27 & 87.78 & 50.21 & 42.8\% & 68.61 &
        37.10 & 37.60 & 14.02 & 57.51 & 17.54 & 69.5\% & 37.25\\
        BBN~\cite{zhou2020bbn} & 
        77.25 & 77.46 &  4.43 & 85.36 & 71.58 & 16.1\% & 77.44 &
        40.02 & 38.96 & 11.19 & 50.69 & 19.16 & 62.2\% & 39.00 \\
        MetaS~\cite{ren2020balanced} & 
        77.55 & 77.24 & 10.28 & \red{92.78} & 61.38 & 33.8\% & 77.30 &
        48.42 & 48.80 & 11.85 & 64.65 & 31.65 & 51.0\% & \green{49.07}\\
        DecTDE~\cite{tang2020long} & 
        79.21 & 79.56 &  \green{3.28} & 85.53 & \blue{76.32} & \green{10.8\%} & 79.24 &
        43.93 & 44.08 & 15.33 & \blue{64.66} & 21.00 & 67.5\% & 43.63\\
        RIDE~\cite{wang2020long} & 
        79.68 & 80.15 &  \blue{3.94} & 87.85 & 75.87 & \blue{13.6\%} & 80.19 &
        42.35 & 42.85 & 17.68 & \red{67.19} & 17.37 & 74.1\% & 42.53\\
        IBLLoss~\cite{park2021influence} & 
        73.33 & 73.35 & 10.24 & 89.82 & 58.72 & 34.6\% & 73.07 &
        38.08 & 38.48 & 13.99 & 57.97 & 18.25 & 68.5\% & 38.46\\
        TADE~\cite{zhang2021test} & 
        80.46 & 80.86 &  4.39 & 89.06 & 76.13 & 14.5\% & \blue{80.93} &
        \green{48.81} & \blue{48.85} &  \blue{9.12} & 60.72 & \blue{34.85} & 42.6\% & \blue{48.63}\\
        LADE~\cite{hong2021disentangling} & 
        \green{82.07} & \green{83.69} &  4.11 & \green{90.05} & \green{77.71} & 13.7\% & 79.81 &
        \blue{48.44} & \green{49.53} &  \green{8.82} & 63.65 & \red{39.38} & \green{38.1\%} & 43.60\\
        Prior-LT~\cite{Xu2021TowardsCM} & 
        71.12 & 72.29 &  4.49 & 81.15 & 66.03 & 18.6\% & 72.51 &
        46.74 & 46.41 &  \red{7.36} & 55.47 & 34.66 & \red{37.5\%} & 46.26 \\
        PCL~\cite{cui2021parametric} & 
        \red{84.10} & \red{84.69} &  \red{2.66} & 89.70 & \red{82.07} &  \red{8.5\%} & \red{84.42} &
        \red{50.78} & \red{50.90} &  9.13 & 62.71 & \green{36.98} & \blue{41.0\%} & \red{50.49} \\
        GCLLoss~\cite{li2022long} & 
        \blue{81.35} & \blue{81.52} &  4.78 & 89.47 & 74.64 & 16.6\% & \green{81.77} &
        47.82 & 47.83 & 14.10 & \green{66.17} & 26.15 & 60.5\% & 47.23 \\
        \hline
      \end{tabular}
      }
  \end{table*}

In real-world applications, limited training data can not reflect the actual data distribution. To evaluate LLTD algorithms more comprehensively, we set up testing datasets which have dynamic evolving distributions. 
The percent of the $c$-th class's samples namely $q_c^{tst}$ is determined according to the following formulation,
\begin{equation}
q_c^{tst}(\alpha) = \frac{({\rho^{tst}})^{-\frac{|c-\alpha|}{C-1}}}{\sum_{c=1}^C ({\rho^{tst}})^{-\frac{|c-\alpha|}{C-1}}},
\end{equation}
where $\alpha$ is a variable controlling the peak of the testing data distribution. Varying $\alpha$ can derive testing data with diversified distributions. 

Suppose the maximum class sample size be $N_{max}^{tst}$. We choose the the total number of samples namely $N^{tst}$ as below,
\begin{equation}
N^{tst} = \sum_{c=1}^C N_{max}^{tst} ({\rho^{tst}})^{-\frac{|c-1|}{C-1}}.
\end{equation}

Then, the sample size of the $c$-th class in testing data can be obtained as, $N_c^{tst}=N^{tst} q_c^{tst}(\alpha)$. Finally, new testing samples can be randomly drawn out from the original dataset.

  \begin{table*}[t]
      \caption{Peformance of existing methods on CIFAR10 and CIFAR100 with $\rho^{trn}=\rho^{tst}=0.05$. 
      }\vspace{-5pt}
      \label{tab:CIFAR-05}
      \fontsize{7.5}{9} \selectfont
      \centering
      \setlength{\tabcolsep}{1.5mm}{
      \begin{tabular}{l|c|c|c|c|c|c|c|c|c|c|c|c|c|c}
        \hline
        \multirow{3}{*}{\textbf{Method}} & \multicolumn{7}{c|}{\bf CIFAR10} & \multicolumn{7}{c}{\bf CIFAR100} \\ \cline{2-15}
        & \multirow{2}{*}{\textbf{AUC}$\uparrow$} & \multicolumn{6}{c|}{\textbf{ACC}} & \multirow{2}{*}{\textbf{AUC}$\uparrow$} & \multicolumn{6}{c}{\textbf{ACC}} \\ \cline{3-8} \cline{10-15}
        & & AVG$\uparrow$ & STD$\downarrow$ & MAX$\uparrow$ & MIN$\uparrow$ & DR$\downarrow$ & BTD$\uparrow$ & & AVG$\uparrow$ & STD$\downarrow$ & MAX$\uparrow$ & MIN$\uparrow$ & DR$\downarrow$ & BTD$\uparrow$ \\ 
        \hline
        CE & 
        80.00 & 80.93 & 4.39 & 88.63 & 75.11 & 15.3\% & 81.22 &
        46.40 & 48.01 & 8.46 & 60.06 & 35.36 & 41.1\% & 47.75 \\
        Focal~\cite{lin2017focal}  & 
        81.49 & 82.38 & 4.35 & 89.80 & 76.47 & 14.8\% & 82.73 &
        47.50 & 49.26 & 9.03 & 61.82 & 36.23 & 41.4\% & 49.18  \\
        LDAM~\cite{cao2019learning} & 
        83.37 & 84.07 & 2.00 & 88.32 & 81.91 & 7.3\% & 84.62 &
        50.79 & 51.98 & 7.16 & 61.36 & 41.07 & 33.1\% & 52.08 \\
        cRT~\cite{kang2019decoupling} & 
        77.65 & 78.64 & 4.21 & 86.29 & 73.50 & 14.8\% & 79.05 &
        46.59 & 48.07 & 7.30 & 57.95 & 37.88 & 34.6\% & 48.21 \\
        BBN~\cite{zhou2020bbn} & 
        84.58 & 85.09 & 1.53 & 88.38 & 83.73 & 5.3\% & 85.19 &
        51.64 & 51.88 & 4.70 & 56.56 & 43.54 & 23.0\% & 51.70 \\
        MetaS~\cite{ren2020balanced} & 
        85.21 & 86.04 & 2.38 & \green{90.74} & 83.28 & 8.2\% & 86.56 &
        \blue{57.56} & \blue{58.53} & 5.11 & \green{65.10} & 51.28 & 21.2\% & \blue{58.76} \\
        DecTDE~\cite{tang2020long} & 
        86.16 & 86.48 & 1.20 & 87.91 & 84.38 & 4.0\% & 86.64 &
        52.25 & 53.20 & 5.75 & 60.80 & 44.50 & 26.8\% & 53.34 \\
        RIDE~\cite{wang2020long} & 
        84.51 & 84.95 & 1.16 & 86.72 & 83.41 & \blue{3.8\%} & 84.76 &
        53.50 & 54.70 & 7.00 & 63.82 & 44.22 & 30.7\% & 54.65 \\
        IBLLoss~\cite{park2021influence} & 
        83.64 & 84.43 & 3.05 & \blue{90.32} & 80.95 & 10.4\% & 84.84 &
        49.82 & 50.93 & 5.73 & 58.59 & 42.84 & 26.9\% & 51.10 \\
        TADE~\cite{zhang2021test} & 
        \blue{86.70} & \blue{87.10} & \blue{1.07} & 89.51 & \blue{85.66} & 4.3\% & \blue{87.25} &
        \green{58.16} & \green{58.85} & \blue{2.97} & 63.16 & \green{54.73} & \blue{13.3\%} & \green{59.19} \\
        LADE~\cite{hong2021disentangling} & 
        86.30 & 87.04 & 1.96 & 89.74 & 84.18 & 6.2\% & 85.97 &
        56.47 & 57.92 & 3.88 & \blue{64.42} & \blue{53.94} & 16.3\% & 55.12 \\
        Prior-LT~\cite{Xu2021TowardsCM} & 
        86.11 & 86.40 & 1.74 & 88.97 & 83.50 & 6.1\% & 86.65 &
        54.79 & 55.16 & \red{2.27} & 58.05 & 51.73 & \red{10.9\%} & 55.21 \\
        PCL~\cite{cui2021parametric} & 
        \red{89.64} & \red{89.88} & \green{1.03} & \red{91.48} & \red{88.08} & \green{3.7\%} & \red{90.14} &
        \red{60.36} & \red{61.05} & \green{2.91} & \red{65.31} & \red{57.18} & \green{12.4\%} & \red{61.52} \\
        GCLLoss~\cite{li2022long} & 
        \green{87.32} & \green{87.66} & \red{1.02} & 89.05 & \green{85.91} & \red{3.5\%} & \green{87.75} &
        57.39 & 57.82 & 3.99 & 62.39 & 51.07 & 18.1\% & 57.77 \\
        \hline
      \end{tabular}
      }
  \end{table*}

Accuracy is the basic metric for evaluating performance of image classification models. 
Intrinsically, it records the percent of correctly predicted testing samples as in the following equation,
\begin{equation}
V^{acc} = \frac{\sum_{n=1}^{N^{tst}} (\hat y_n = y_n) }{N^{tst}}.
\end{equation} 

We synthesize a series of testing sets by varying $\alpha$ in $\{\frac{(t-1)C}{T}+1\}_{t=1}^T$, where $T$ denotes the times of data synthesization. 
For each times of synthesization, we implement five replaceable samplings.
The overall performance on this testing set synthesization can be evaluated by averaging the five samplings' accuracy, which is defined as $V^{acc}_{t}$.
The distribution shift between the training set and the $t$-th synthesized testing set is estimated as,
\begin{equation}
    \delta_t = -\sum_{c=1}^C [q_c^{trn} \ln(\frac{q_c^{trn}}{q_c^{tst}(\alpha_t) })+q_c^{tst}(\alpha_t) \ln(\frac{q_c^{tst}(\alpha_t)}{q_c^{trn}})],
\end{equation}
where $\alpha_t=\frac{(t-1)C}{T}+1$.

One simple manner to unify the results of $T$ synthesizations is directly averaging them:
\begin{equation}
V^{avg} = \frac{1}{T}\sum_{t=1}^T V^{acc}_{t},
\end{equation}
where $V^{avg}$ denotes the averaged accuracy. We can also evaluate the sensitivity to distribution variation with the accuracy drop ratio $V^{dr}$ and the standard deviation $V^{std}$:
\begin{align}
V^{dr}  & = \frac{\max_{t} V^{acc}_{t} - \min_{t} V^{acc}_{t}}{\max_{t} V^{acc}_{t}}, \\
V^{std} & = \sqrt{\sum_{t=1}^T (V^{acc}_{t} - V^{avg})^2/T}.
\end{align}

  \begin{table*}[t]
      \fontsize{7.5}{9} \selectfont
      \centering
      \caption{Peformance of existing methods on CIFAR10 and CIFAR100 with $\rho^{trn}=\rho^{tst}=0.1$. 
      }\vspace{-5pt}
      \label{tab:CIFAR-1}
      \setlength{\tabcolsep}{1.5mm}{
      \begin{tabular}{l|c|c|c|c|c|c|c|c|c|c|c|c|c|c}
        \hline
        \multirow{3}{*}{\textbf{Method}} & \multicolumn{7}{c|}{\bf CIFAR10} & \multicolumn{7}{c}{\bf CIFAR100} \\ \cline{2-15}
        & \multirow{2}{*}{\textbf{AUC}$\uparrow$} & \multicolumn{6}{c|}{\textbf{ACC}} & \multirow{2}{*}{\textbf{AUC}$\uparrow$} & \multicolumn{6}{c}{\textbf{ACC}} \\ \cline{3-8} \cline{10-15}
        & & AVG$\uparrow$ & STD$\downarrow$ & MAX$\uparrow$ & MIN$\uparrow$ & DR$\downarrow$ & BTD$\uparrow$ & & AVG$\uparrow$ & STD$\downarrow$ & MAX$\uparrow$ & MIN$\uparrow$ & DR$\downarrow$ & BTD$\uparrow$ \\ 
        \hline
        CE & 
        83.83 & 84.53 & 2.17 & 88.63 & 82.23 & 7.2\% & 84.38 &
        51.93 & 53.2 & 5.47 & 60.81 & 45.0 & 26.0\% & 53.01 \\
        Focal~\cite{lin2017focal}  & 
        84.32 & 84.91 & 1.98 & 88.51 & 82.61 & 6.7\% & 85.09 &
        53.43 & 54.61 & 5.34 & 61.87 & 46.44 & 24.9\% & 54.29 \\
        LDAM~\cite{cao2019learning} & 
        86.4 & 86.68 & \blue{0.79} & 88.48 & 85.75 & 3.1\% & 86.69 &
        55.06 & 55.96 & 4.47 & 62.18 & 48.79 & 21.5\% & 55.26 \\
        cRT~\cite{kang2019decoupling} & 
        83.15 & 83.82 & 1.91 & 87.45 & 82.03 & 6.2\% & 84.09 &
        51.45 & 52.48 & 4.34 & 58.54 & 45.84 & 21.7\% & 52.29 \\
        BBN~\cite{zhou2020bbn} & 
        86.85 & 87.02 & \green{0.55} & 87.67 & 86.04 & \green{1.9\%} & 87.06 &
        57.3 & 57.38 & 1.72 & 59.18 & 54.14 & 8.5\% & 57.11 \\
        MetaS~\cite{ren2020balanced} & 
        \blue{89.38} & \green{89.78} & 1.16 & \red{92.2} & \blue{88.6} & 3.9\% & \green{89.8} &
        \green{62.05} & \green{62.66} & 2.61 & \red{66.29} & 58.86 & 11.2\% & \green{62.31}\\
        DecTDE~\cite{tang2020long} & 
        87.39 & 87.37 &  1.21 & 89.21 & 85.64 & 4.0\% & 87.47 &
        57.59 & 58.44 & 4.2 & 63.97 & 51.63 & 19.3\% & 57.83\\
        RIDE~\cite{wang2020long} & 
        85.62 & 85.81 &  0.91 & 87.16 & 84.5 & 3.1\% & 85.98 &
        55.63 & 56.59 & 4.37 & 62.53 & 49.78 & 20.4\% & 56.12\\
        IBLLoss~\cite{park2021influence} & 
        86.67 & 87.07 & 1.32 & 89.56 & 85.48 & 4.6\% & 86.82 &
        53.08 & 53.87 & 3.53 & 58.6 & 48.45 & 17.3\% & 53.66\\
        TADE~\cite{zhang2021test} & 
        \green{89.45} & \blue{89.64} & \red{0.51} & \blue{90.54} & \green{88.97} & \red{1.7\%} & \blue{89.76} &
        \blue{61.32} & \blue{61.67} & \blue{1.32} & 63.8 & \green{59.74} & 7.5\% & \blue{61.36}\\
        LADE~\cite{hong2021disentangling} & 
        88.49 & 88.78 & 1.07 & 90.3 & 87.25 & 3.4\% & 88.47 &
        60.28 & 61.13 & 2.16 & \blue{65.15} & \blue{59.03} & 9.4\% & 58.91\\
        Prior-LT~\cite{Xu2021TowardsCM} & 
        87.78 & 87.8 & 1.05 & 89.47 & 86.42 & 3.4\% & 87.79 &
        57.75 & 57.91 & \green{0.97} & 59.32 & 56.18 & \green{5.3\%} & 57.73 \\
        PCL~\cite{cui2021parametric} & 
        \red{90.77} & \red{90.86} & \blue{0.79} & \green{92.02} & \red{89.58} & 2.7\% & \red{91.25} &
        \red{63.39} & \red{63.82} & 1.73 & \green{66.19} & \red{61.33} & \blue{7.3\%} & \red{63.81} \\
        GCLLoss~\cite{li2022long} & 
        89.2 & 89.23 & 0.86 & 90.47 & 88.18 & \blue{2.5\%} & 89.45 &
        60.74 & 60.65 & \red{0.88} & 61.56 & 58.68 & \red{4.7\%} & 60.24 \\
        \hline
      \end{tabular}
      }
  \end{table*}

The other manner for combining the accuracy values is calculating the area under the curve of $T$ synthesizations' results. All testing sets are ranked with respect to the distribution shifts between training and testing sets in the ascending order. Denote the ranked set indices as $\{t_k\}_{k=1}^T$. This metric is calculated as below:
\begin{equation}
V^{auc} = \frac{\sum_{k=1}^{T-1} (V^{acc}_{t_k} + V^{acc}_{t_{k+1}})(\delta_{t_{k+1}} - \delta_{t_{k}})/2}{ \delta_{t_{T}} - \delta_{t_{1}} }.
\end{equation}


\section{Benchmarks}
\label{sec:exper}

\subsection{Experimental settings.}
We use two datasets for training and testing, including CIFAR10 and CIFAR100~\cite{krizhevsky2009learning}. 
Both CIFAR10 and CIFAR100 contain 60,000 images with size of 32$\times$32. 
The number of classes of CIFAR10 and CIFAR100 is 10 and 100, respectively. 
The original datasets are split into 50,000 images for training and 10,000 images for testing.
We follow~\cite{kang2019decoupling,zhou2020bbn} to synthesize imbalanced training set with the imbalance ratio $\rho^{trn}\in\{0.01,0.05,0.1\}$. 
$N_c^{trn}$ is set as 5,000 and 500 for CIFAR10 and CIFAR100, respectively.
During training, 10\% images of each class are used for validation.
When generating testing data, we choose $\rho^{tst}$ from $\{ 0.01, 0.05, 0.1\}$ and set $N_{max}^{tst}$ as 1,000 and 100 for CIFAR10 and CIFA100, respectively.

\subsection{Benchmark 1: Testing Data having Same Imbalance Ratio as Training Data} \label{sec:exper-same}
In this subsection, we re-evaluate the performance of existing methods, including Focal~\cite{lin2017focal}, LDAM~\cite{cao2019learning}, cRT~\cite{kang2019decoupling}, BBN~\cite{zhou2020bbn}, MetaS~\cite{ren2020balanced}, DecTDE~\cite{tang2020long}, RIDE~\cite{wang2020long}, IBLoss~\cite{park2021influence}, TADE~\cite{zhang2021test}, LADE~\cite{hong2021disentangling}, Prior-LT~\cite{Xu2021TowardsCM}, PCL~\cite{cui2021parametric}, and GCLLoss~\cite{li2022long}. 
The baseline method (CE) is implemented using the conventional uniform data sampling and standard cross-entropy loss function.
For all methods, we use ResNet32~\cite{he2016deep} as the classification backbone model.
Here, we use the same imbalance ratio for training and testing data.

\begin{figure*}[t]
    \centering
    \begin{subfigure}{0.328\linewidth}
        \centering
        \includegraphics[width=1\linewidth, trim=10 10 5 5, clip]{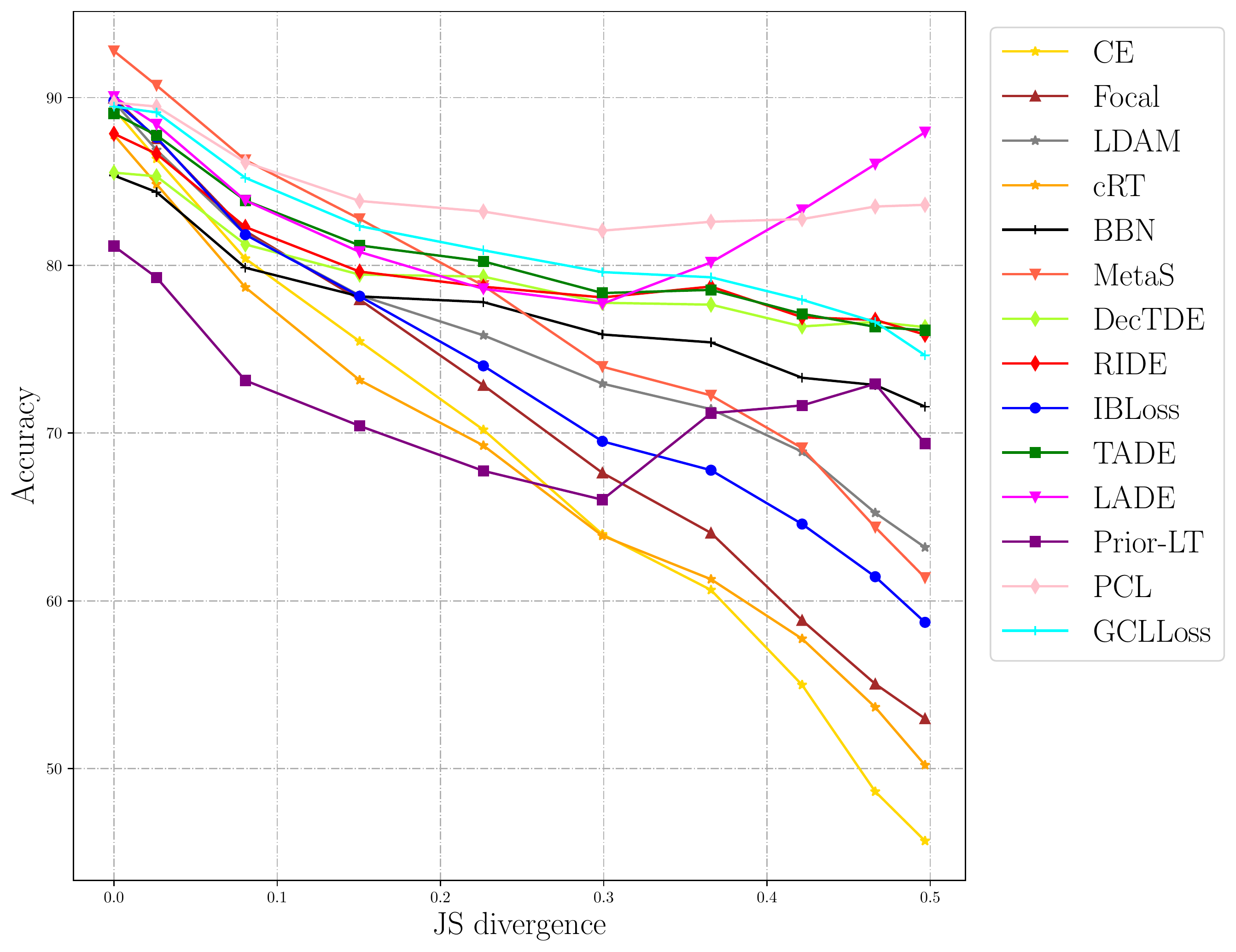}
        \caption{$\rho^{trn}=\rho^{tst}=0.01$}
    \end{subfigure}
  \begin{subfigure}{0.328\linewidth}
        \centering
        \includegraphics[width=1\linewidth,trim=10 10 5 5, clip]{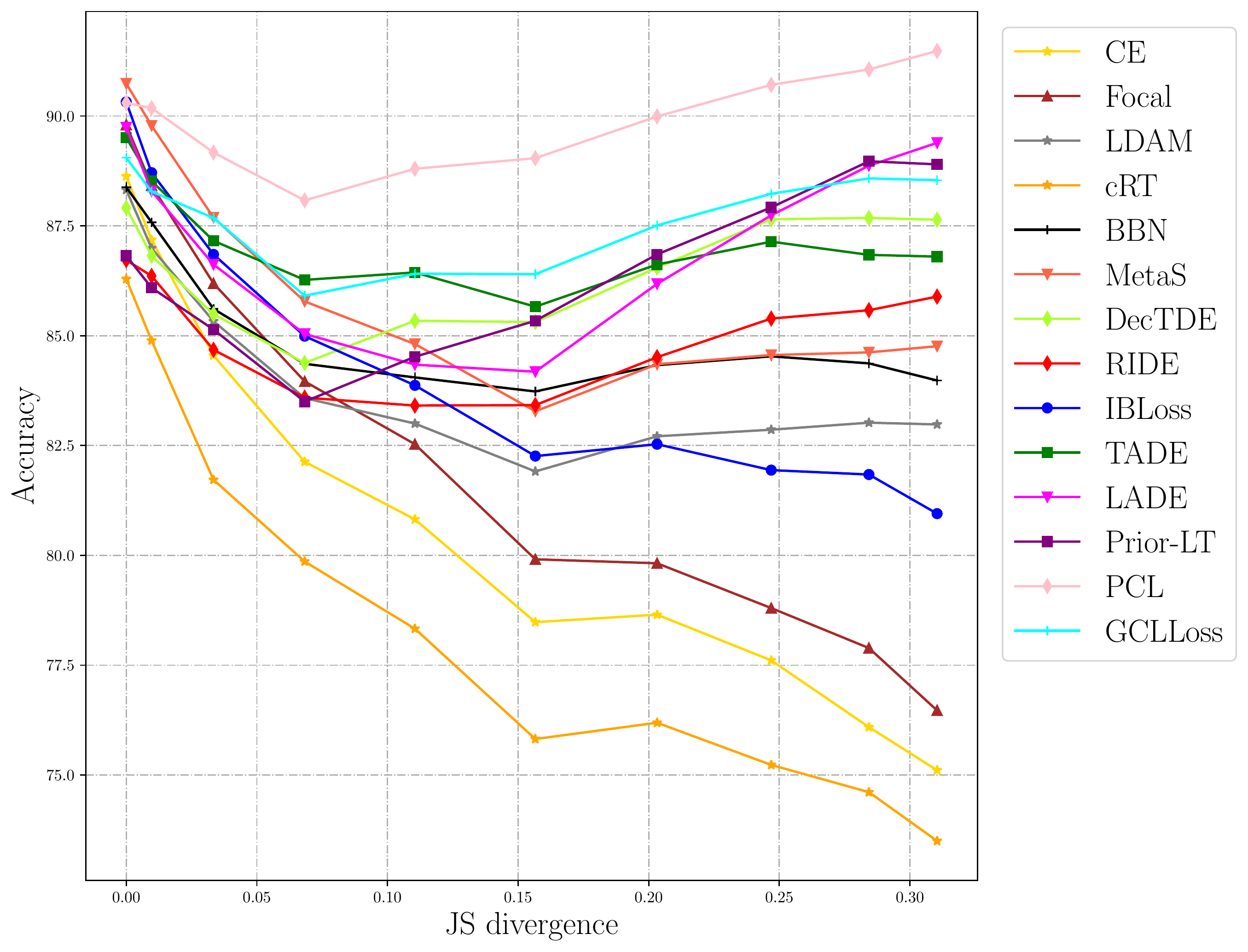}
    \caption{$\rho^{trn}=\rho^{tst}=0.05$}
    \end{subfigure}
  \begin{subfigure}{0.328\linewidth}
        \centering
        \includegraphics[width=1\linewidth, trim=10 10 5 5, clip]{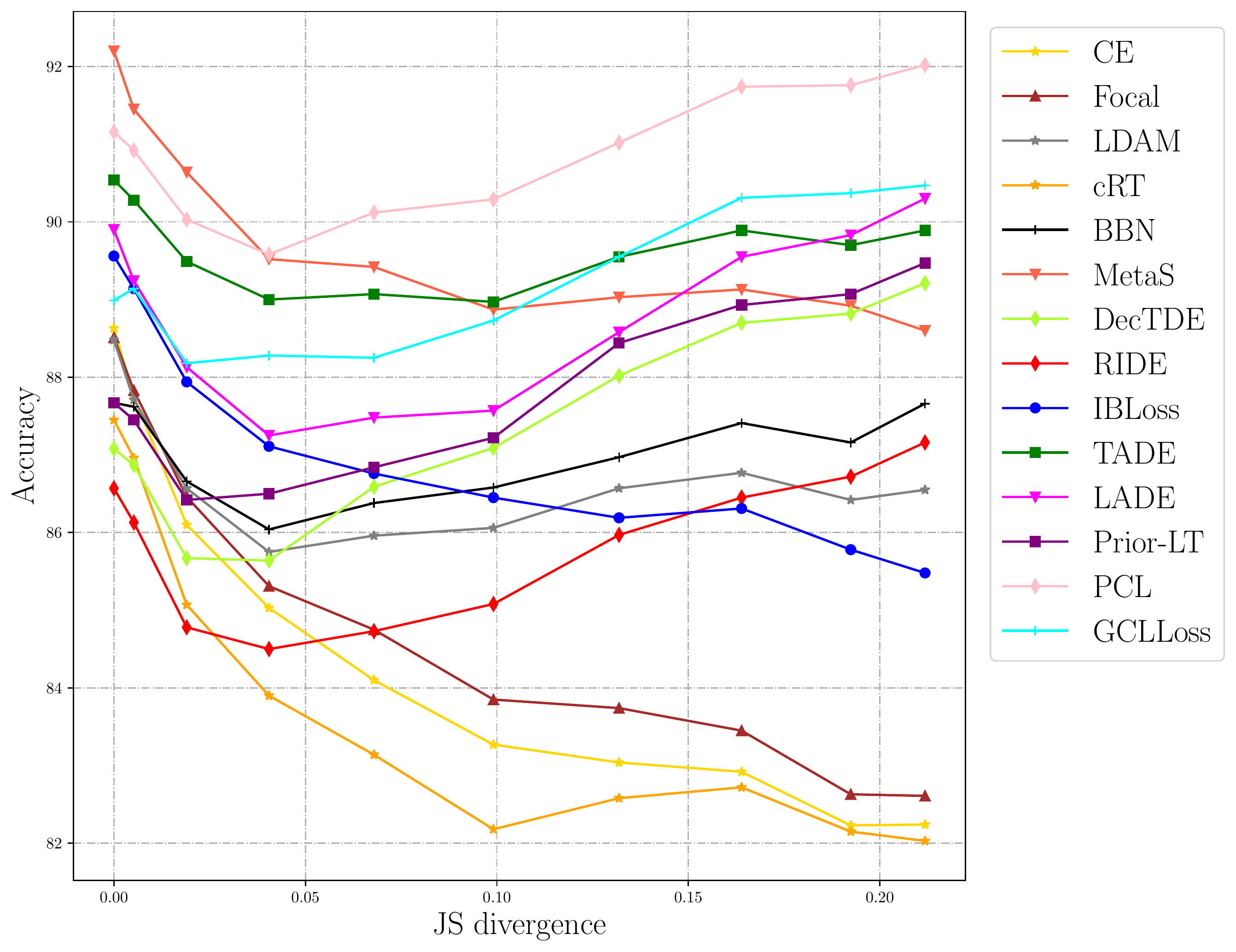}
        \caption{$\rho^{trn}=\rho^{tst}=0.1$}
    \end{subfigure}
\caption{Accuracy curves of existing methods on CIFAR10 against the JS divergence. (a) The decay coefficients $\rho^{trn}$ and $\rho^{tst}$ are set to 0.01; (b) $\rho^{trn}$ and $\rho^{tst}$ are set to 0.05; (c) $\rho^{trn}$ and $\rho^{tst}$ are set to 0.1.}
\label{fig:acc-curve-cifar10}
\end{figure*}

\begin{figure*}[t]
    \centering
    \begin{subfigure}{0.328\linewidth}
        \centering
        \includegraphics[width=1\linewidth, trim=10 10 5 5, clip]{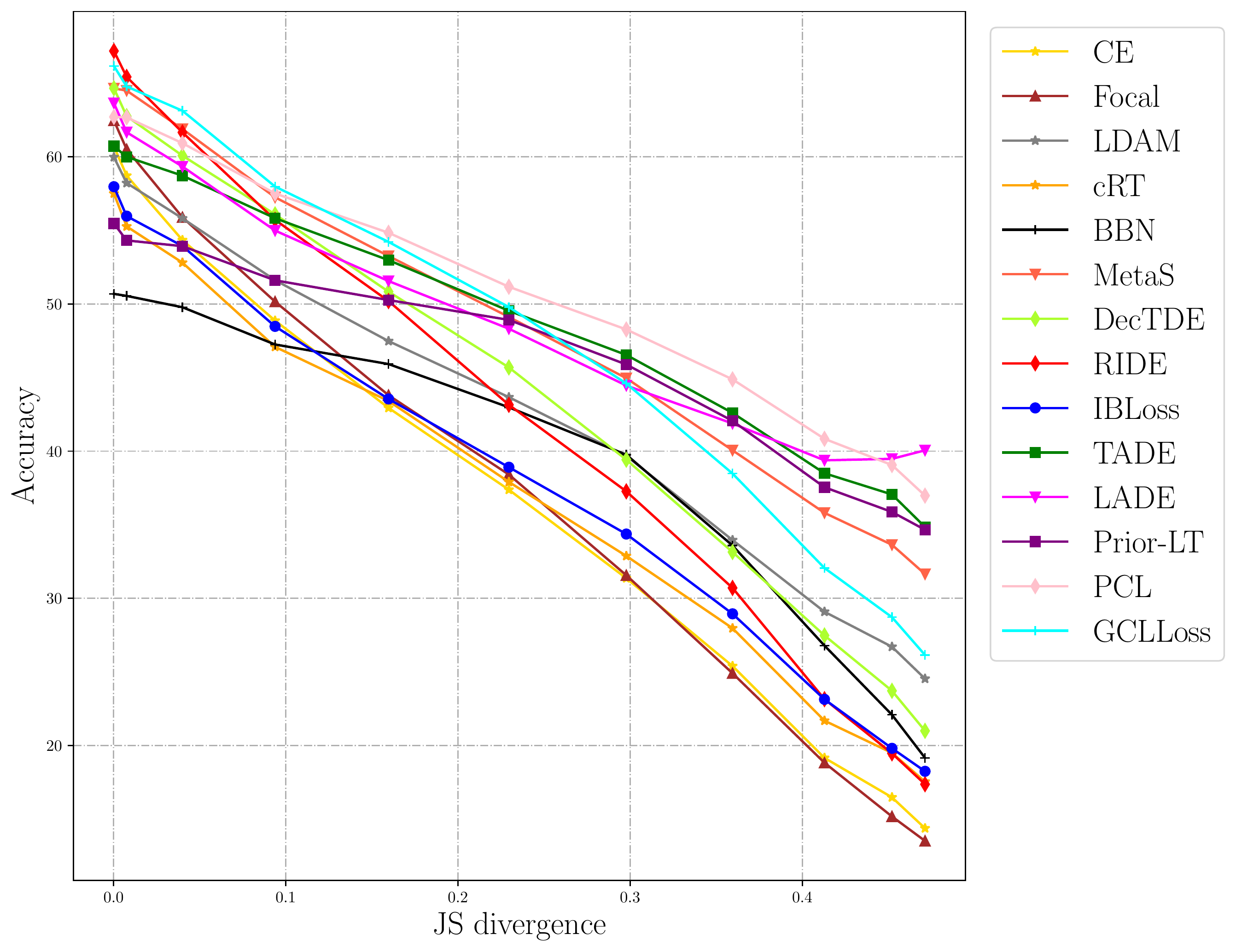}
        \caption{$\rho^{trn}=\rho^{tst}=0.01$}
    \end{subfigure}
  \begin{subfigure}{0.328\linewidth}
        \centering
        \includegraphics[width=1\linewidth,trim=10 10 5 5, clip]{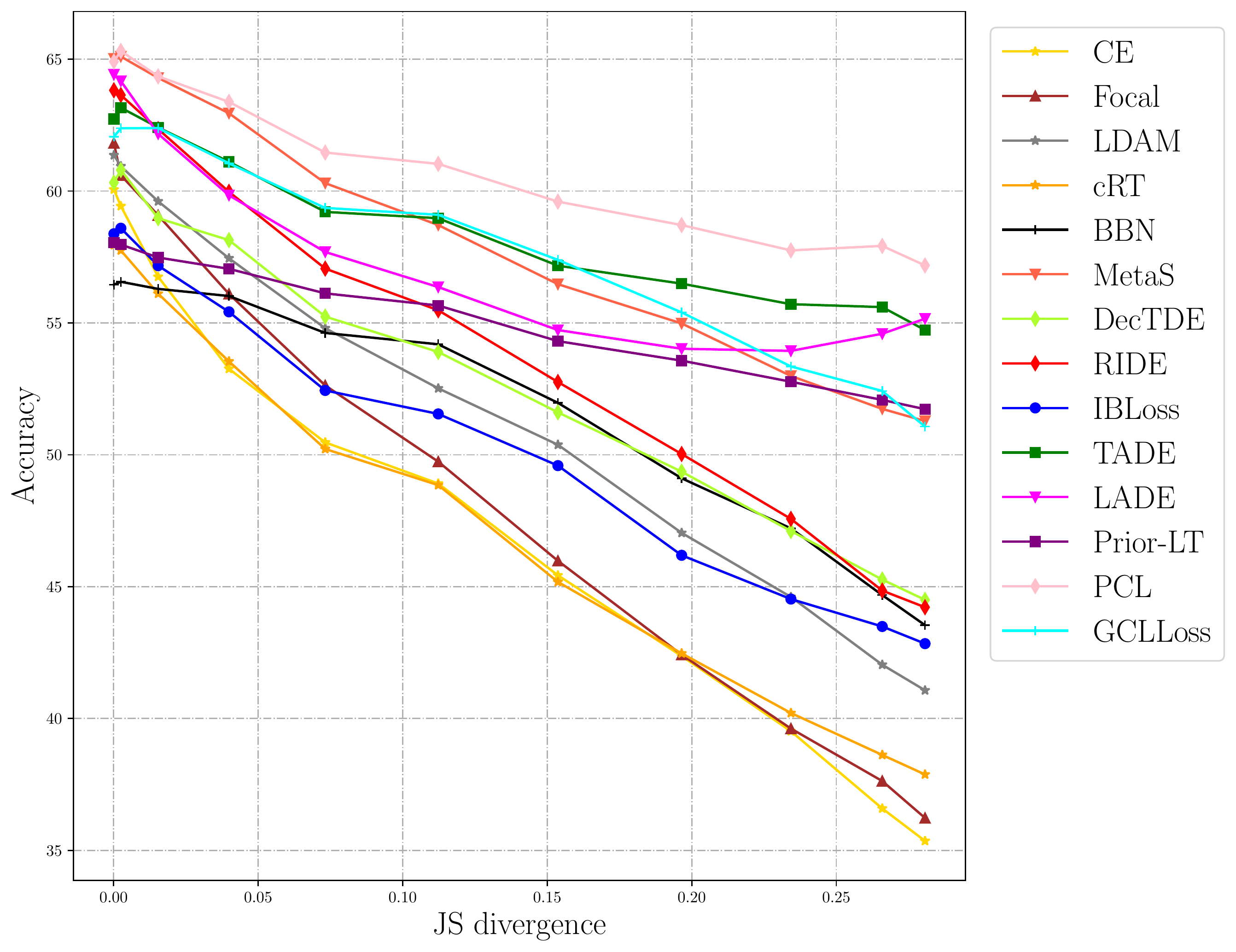}
        \caption{$\rho^{trn}=\rho^{tst}=0.05$}
    \end{subfigure}
  \begin{subfigure}{0.328\linewidth}
        \centering
        \includegraphics[width=1\linewidth, trim=10 10 5 5, clip]{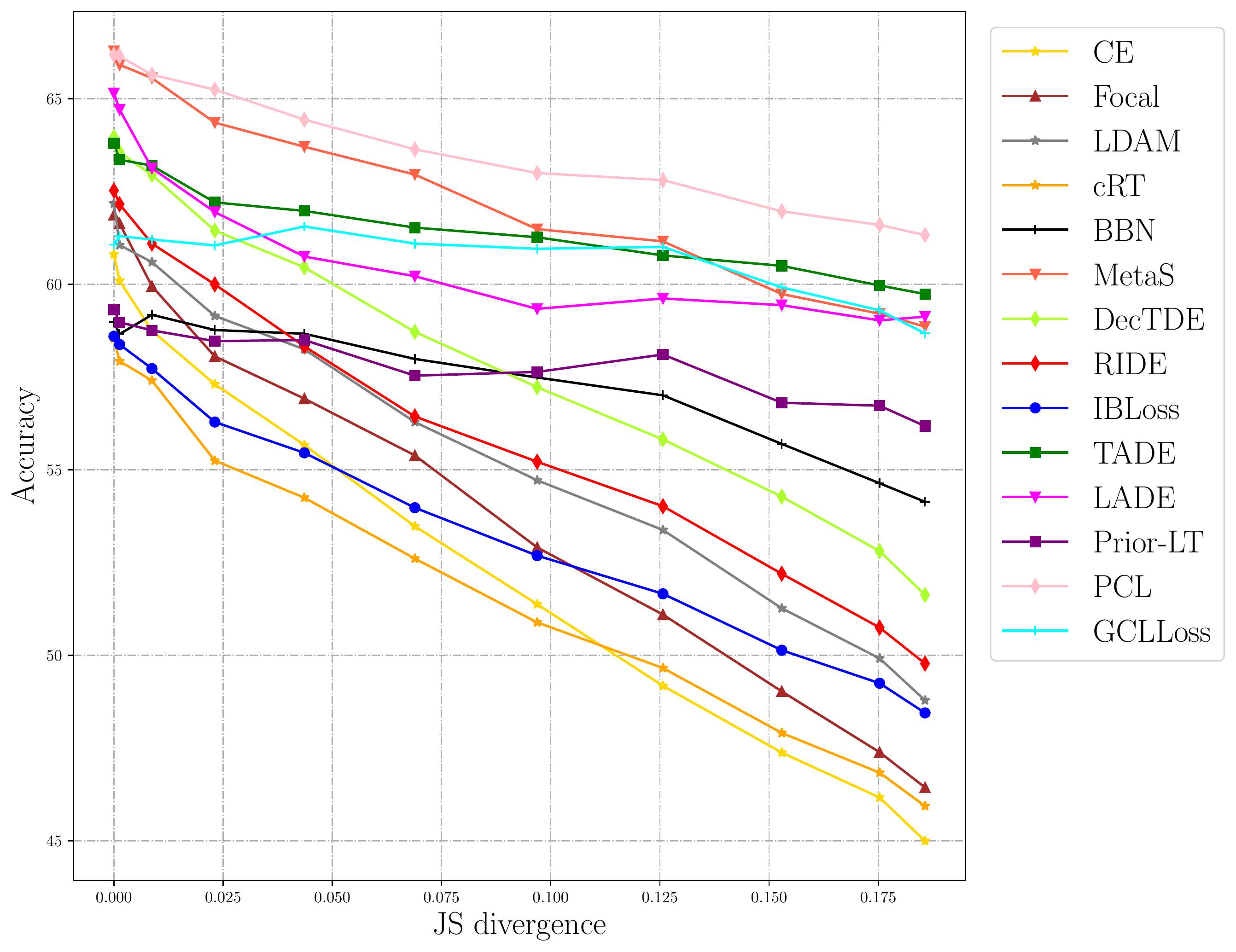}
        \caption{$\rho^{trn}=\rho^{tst}=0.1$}
    \end{subfigure}
\caption{Accuracy curves of existing methods on CIFAR100 against the JS divergence.}
\label{fig:acc-curve-cifar100}
\end{figure*}

\begin{table*}[t]
      \fontsize{7.5}{9} \selectfont
      \centering
      \caption{Performance of existing methods on CIFAR10 and CIFAR100 with $\rho^{trn}=0.05$ and $\rho^{tst}=0.01$. 
      } \vspace{-5pt}
      \label{tab:CIFAR-05-01}
      \setlength{\tabcolsep}{1.5mm}{
      \begin{tabular}{l|c|c|c|c|c|c|c|c|c|c|c|c|c|c}
        \hline
        \multirow{3}{*}{\textbf{Method}} & \multicolumn{7}{c|}{\bf CIFAR10} & \multicolumn{7}{c}{\bf CIFAR100} \\ \cline{2-15}
        & \multirow{2}{*}{\textbf{AUC}$\uparrow$} & \multicolumn{6}{c|}{\textbf{ACC}} & \multirow{2}{*}{\textbf{AUC}$\uparrow$} & \multicolumn{6}{c}{\textbf{ACC}} \\ \cline{3-8} \cline{10-15}
        & & AVG$\uparrow$ & STD$\downarrow$ & MAX$\uparrow$ & MIN$\uparrow$ & DR$\downarrow$ & BTD$\uparrow$ & & AVG$\uparrow$ & STD$\downarrow$ & MAX$\uparrow$ & MIN$\uparrow$ & DR$\downarrow$ & BTD$\uparrow$ \\ 
        \hline
        CE & 
        79.18 & 80.96 & 5.51 & 91.26 & 74.05 & 18.9\% & 81.22 & 
        45.10 & 47.92 & 11.08 & 63.03 & 40.99 & 35.0\% & 47.75 \\
        Focal~\cite{lin2017focal}  & 
        80.43 & 82.27 & 5.69 & \bcg{92.24} & 75.33 & 18.3\% & 82.73 & 
        46.30 & 49.51 & 12.02 & 65.47 & 47.91 & 26.8\% & 49.18 \\
        LDAM~\cite{cao2019learning} & 
        82.82 & 84.11 & 2.85 & 90.17 & 81.81 & 9.3\% & 84.62 & 
        49.87 & 51.98 & 9.53 & 63.80 & \bcg{54.91} & \bcb{13.9\%} & 52.08 \\
        cRT~\cite{kang2019decoupling} & 
        76.63 & 78.67 & 5.63 & 88.92 & 72.68 & 18.3\% & 79.05 & 
        45.51 & 48.10 & 9.26 & 61.15 & \bcb{53.55} & \bcg{12.4\%} & 48.21 \\
        BBN~\cite{zhou2020bbn} & 
        84.14 & 85.05 & 2.10 & 89.47 & 83.50 & 6.7\% & 85.19 & 
        51.61 & 51.94 & 6.59 & 57.98 & 49.66 & 14.3\% & 51.70 \\
        MetaS~\cite{ren2020balanced} & 
        84.68 & 86.15 & 3.21 & \bcr{92.27} & 83.03 & 10.0\% & 86.56 & 
        57.08 & 58.76 & 6.90 & \bcg{67.53} & 31.89 & 52.8\% & \bcb{58.76} \\
        DecTDE~\cite{tang2020long} & 
        85.93 & 86.44 & 1.86 & 88.92 & 82.93 & 6.7\% & 86.64 & 
        51.57 & 54.12 & 7.85 & 63.05 & 30.92 & 51.0\% & 53.34 \\
        RIDE~\cite{wang2020long} & 
        84.09 & 84.84 & 1.82 & 87.48 & 82.07 & 6.2\% & 84.76 & 
        52.24 & 54.52 & 9.35 & \bcb{66.96} & 44.40 & 33.7\% & 54.65 \\
        IBLLoss~\cite{park2021influence} & 
        82.92 & 84.44 & 4.00 & 92.11 & 80.75 & 12.3\% & 84.84 & 
        49.17 & 51.03 & 7.41 & 60.94 & \bcr{55.72} & \bcr{8.6\%} & 51.10 \\
        TADE~\cite{zhang2021test} & 
        86.31 & 87.03 & \bcg{1.63} & 90.54 & \bcg{85.53} & \bcb{5.5\%} & \bcb{87.25} & 
        \bcb{57.83} & \bcb{58.88} & \bcg{3.69} & 63.72 & 34.72 & 45.5\% & \bcg{59.19} \\
        LADE~\cite{hong2021disentangling} & 
        \bcg{87.22} & \bcg{88.39} & 2.94 & \bcb{92.17} & 84.48 & 8.3\% & 85.97 & 
        \bcg{58.11} & \bcg{60.63} & 4.99 & \bcr{69.48} & 36.40 & 47.6\% & 55.12 \\
        Prior-LT~\cite{Xu2021TowardsCM} & 
        85.90 & 86.41 & 2.57 & 89.86 & 81.88 & 8.9\% & 86.65 & 
        54.68 & 55.26 & \bcr{3.18} & 59.22 & 39.21 & 33.8\% & 55.21 \\
        PCL~\cite{cui2021parametric} & 
        \bcr{89.40} & \bcr{89.78} & \bcr{1.56} & 91.91 & \bcr{86.99} & \bcg{5.4\%} & \bcr{90.14} & 
        \bcr{60.09} & \bcr{61.20} & \bcb{3.81} & 66.31 & 39.61 & 40.3\% & \bcr{61.52} \\
        GCLLoss~\cite{li2022long} & 
        \bcb{87.14} & \bcb{87.70} & \bcb{1.68} & 89.55 & \bcb{84.78} & \bcr{5.3\%} & \bcg{87.75} & 
        56.91 & 57.86 & 5.23 & 63.72 & 48.99 & 23.1\% & 57.77 \\
        \hline
      \end{tabular}
      }
  \end{table*}

We report the experimental results on re-distributed versions of CIFAR10 and CIFAR100 which have the same imbalance ratios for training and testing data in Table~\ref{tab:CIFAR-01} ($\rho^{trn}=\rho^{tst}=0.01$), Table~\ref{tab:CIFAR-05} ($\rho^{trn}=\rho^{tst}=0.05$), and Table~\ref{tab:CIFAR-1} ($\rho^{trn}=\rho^{tst}=0.1$). The accuracy curves with respect to the JS divergence are illustrated in Fig.~\ref{fig:acc-curve-cifar10} and Fig.~\ref{fig:acc-curve-cifar100}.

\begin{table*}[t]
      \fontsize{7.5}{9} \selectfont
      \centering
      \caption{Peformance of existing methods on CIFAR10 and CIFAR100 with $\rho^{trn}=0.05$ and $\rho^{tst}=0.1$. 
      }\vspace{-5pt}
      \label{tab:CIFAR-05-1}
      
      \setlength{\tabcolsep}{1.5mm}{
      \begin{tabular}{l|c|c|c|c|c|c|c|c|c|c|c|c|c|c}
        \hline
        \multirow{3}{*}{\textbf{Method}} & \multicolumn{7}{c|}{\bf CIFAR10} & \multicolumn{7}{c}{\bf CIFAR100} \\ \cline{2-15}
        & \multirow{2}{*}{\textbf{AUC}$\uparrow$} & \multicolumn{6}{c|}{\textbf{ACC}} & \multirow{2}{*}{\textbf{AUC}$\uparrow$} & \multicolumn{6}{c}{\textbf{ACC}} \\ \cline{3-8} \cline{10-15}
        & & AVG$\uparrow$ & STD$\downarrow$ & MAX$\uparrow$ & MIN$\uparrow$ & DR$\downarrow$ & BTD$\uparrow$ & & AVG$\uparrow$ & STD$\downarrow$ & MAX$\uparrow$ & MIN$\uparrow$ & DR$\downarrow$ & BTD$\uparrow$ \\ 
        \hline
        CE & 
        80.48 & 80.97 & 3.30 & 86.61 & 76.59 & 11.6\% & 81.22 & 
        47.06 & 47.87 & 6.76 & 57.26 & 37.83 & 33.9\% & 47.75 \\
        Focal~\cite{lin2017focal}  & 
        81.91 & 82.40 & 3.37 & 88.00 & 77.84 & 11.5\% & 82.73 & 
        48.38 & 49.24 & 7.33 & 59.39 & 38.53 & 35.1\% & 49.18 \\
        LDAM-DRW~\cite{cao2019learning} & 
        83.80 & 84.17 & 1.44 & 87.15 & 82.90 & 4.9\% & 84.62 & 
        51.44 & 51.96 & 5.79 & 59.40 & 43.05 & 27.5\% & 52.08 \\
        cRT~\cite{kang2019decoupling} & 
        78.05 & 78.64 & 3.35 & 84.43 & 74.59 & 11.7\% & 79.05 & 
        47.44 & 48.14 & 5.62 & 55.95 & 40.04 & 28.4\% & 48.21 \\
        BBN~\cite{zhou2020bbn} & 
        84.79 & 85.07 & 1.09 & 87.32 & 84.05 & 3.7\% & 85.19 & 
        52.11 & 52.07 & 3.82 & 56.01 & 45.24 & 19.2\% & 51.70 \\
        MetaS~\cite{ren2020balanced} & 
        85.66 & 86.11 & 1.77 & \bcg{89.40} & 84.57 & 5.4\% & 86.56 & 
        \bcb{58.10} & \bcb{58.61} & 4.17 & \bcg{64.38} & 52.39 & 18.6\% & \bcb{58.76}\\
        DecTDE~\cite{tang2020long} & 
        86.23 & 86.41 & 0.81 & 87.37 & 85.02 & 2.7\% & 86.64 & 
        52.76 & 53.17 & 4.56 & 59.24 & 46.11 & 22.2\% & 53.34\\
        RIDE~\cite{wang2020long} & 
        84.61 & 84.86 & 0.75 & 86.02 & 83.80 & 2.6\% & 84.76 & 
        54.01 & 54.60 & 5.66 & 62.31 & 45.79 & 26.5\% & 54.65\\
        IBLLoss~\cite{park2021influence} & 
        84.04 & 84.48 & 2.33 & \bcb{88.71} & 81.73 & 7.9\% & 84.84 & 
        50.28 & 50.81 & 4.57 & 57.18 & 43.97 & 23.1\% & 51.10\\
        TADE~\cite{zhang2021test} & 
        \bcb{86.80} & \bcb{87.03} & \bcr{0.68} & 88.50 & \bcb{86.24} & \bcb{2.6\%} & \bcb{87.25} & 
        \bcg{58.51} & \bcg{58.84} & \bcg{2.38} & \bcb{62.32} & \bcg{55.22} & \bcb{11.4\%} & \bcg{59.19} \\
        LADE~\cite{hong2021disentangling} & 
        85.99 & 86.47 & 1.39 & 88.29 & 84.60 & 4.2\% & 85.97 & 
        55.95 & 56.79 & 3.02 & 62.17 & \bcb{53.64} & 13.7\% & 55.12 \\
        Prior-LT~\cite{Xu2021TowardsCM} & 
        86.32 & 86.47 & 1.24 & 88.27 & 84.70 & 4.0\% & 86.65 & 
        55.01 & 55.18 & \bcr{1.96} & 57.79 & 51.80 & \bcr{10.4\%} & 55.21 \\
        PCL~\cite{cui2021parametric} & 
        \bcr{89.76} & \bcr{89.89} & \bcr{0.68} & \bcr{90.75} & \bcr{88.76} & \bcr{2.2\%} & \bcr{90.14} & 
        \bcr{60.84} & \bcr{61.18} & \bcb{2.40} & \bcr{64.70} & \bcr{57.73} & \bcg{10.8\%} & \bcr{61.52} \\
        GCLLoss~\cite{li2022long} & 
        \bcg{87.41} & \bcg{87.62} & \bcb{0.74} & 88.41 & \bcg{86.46} & \bcr{2.2\%} & \bcg{87.75} & 
        57.69 & 57.84 & 3.24 & 61.89 & 52.28 & 15.5\% & 57.77 \\
        \hline
      \end{tabular}
      }
  \end{table*}

We can observe that, the parametric contrastive learning~\cite{cui2021parametric} is very good at coping with the data imbalance problem. 
It achieves the highest classification accuracy overall, and showcases high stability across different testing distributions. 
This means that learning good representations is a very effective strategy for relieving the influence of data imbalance, due to the improvement in feature generalization ability and the prevention of overfitting with tail classes.
TADE~\cite{zhang2021test} also has very promising ability in tackling the data imbalance problem, due to the combination of multiple expert models learned with different simulated training distributions.
Another effective direction is adjusting the classification logits according to classes' sample sizes like GCLoss~\cite{li2022long} and MetaS~\cite{ren2020balanced}, which can help to enlarge the embedding space of tail classes. 
LADE~\cite{zhang2021test} is targeted at disentangling the label distribution of the training data from the model prediction. It also has fine effect in adapting to arbitrary testing distributions. For example, it produces the second best AUC and average ACC metrics on CIFAR10 with $\rho^{trn}=\rho^{tst}=0.01$.

From the metric values on standard deviation (STD) and drop ratio (DR), most methods with relatively better classification performance such as PCL, TADE, and GCLoss are robust against the variance of testing data distribution. Prior-LT~\cite{Xu2021TowardsCM} which uses class-balanced mixup to construct training samples and prior-compensated Softmax for probability prediction have relatively more stable performance than other methods on CIFA100. 

The upper bound of LLTD algorithms on testing data with unknown distribution can be approximated by the maximum accuracy (ACC-MAX) metric. Most of algorithms are unable to improve this metric. On CIFA100, a few algorithms such as cRT, BBN, IBLoss and Prior-LT  even achieve severely decreased ACC-MAX value compared to the baseline. The reason may be that those methods excessively bias the training process towards tail classes. Several methods are capable of improving the ACC-MAX metric, such as Meta-Softmax, PCL, LADE, focal loss, etc.

The minimum accuracy (ACC-MIN) can approximate the lower bound of LLTD algorithms. This metric is usually obtained when there exists large or moderate shift between training and testing distributions.  We can see that most methods are capable of improving this metric, since they have specific designs for enhancing the significance of tail classes during training. The most three effective methods in boosting ACC-MIN are PCL, TADE, and LADE.

\subsection{Benchmark 2: Testing Data with Imbalance Ratios Different to Training Data} \label{sec:larger}
We report the experimental results on re-distributed versions of CIFAR10 and CIFAR100 where the imbalance ratios of the training data are different with that of the testing data in Table~\ref{tab:CIFAR-05-01} ($\rho^{trn}=0.05$, $\rho^{tst}=0.01$) and Table~\ref{tab:CIFAR-05-1} ($\rho^{trn}=0.05$, $\rho^{tst}=0.1$).
We can see that existing methods have similar performance rankings with the situation where the imbalance ratios of training and testing distributions are same.

\section{Discussions and Conclusions}
\label{sec:conclu}

\textbf{Discussions}. Based on the experimental results, we recommend the following directions to improve algorithms for learning with long-tailed distribution.
\begin{itemize}
  \item The minimum accuracy of most methods is not high under severe data imbalance or large number of classes (e.g., CIFAR100). This indicates that there still exists huge space for improving the performance on tail classes. 
  \item From Fig.~\ref{fig:acc-curve-cifar10}, on CIFAR10, a few methods, such as  PCL, GCLoss, and LADE, are capable of achieving high accuracy on small and large JS divergence. However, the accuracy on middle JS divergence is relatively low. This means that they are able to achieve good performance on head and tail classes but the middle classes lack attention during training.
\end{itemize}

\textbf{Conclusions}. In this paper, we set up new Benchmarks to analyze the performance of methods for learning with long-tailed distribution. Based on a series of testing sets with evolving data distribution, we devise new metrics to analyze the accuracy, stability, and upper/lower bound of existing methods comprehensively. Extensive experiments on CIFAR10 and CIFAR100 are conducted to evaluate existing methods. We also summarize existing methods into data, feature, loss, and prediction balancing types according to the focused stage in the working pipeline.

{\small

\begin{thebibliography}{10}\itemsep=-1pt

\bibitem{alshammari2022long}
Shaden Alshammari, Yu-Xiong Wang, Deva Ramanan, and Shu Kong.
\newblock Long-tailed recognition via weight balancing.
\newblock In {\em CVPR}, pages 6897--6907, 2022.

\bibitem{cai2021ace}
Jiarui Cai, Yizhou Wang, and Jenq-Neng Hwang.
\newblock Ace: Ally complementary experts for solving long-tailed recognition
  in one-shot.
\newblock In {\em Proceedings of the IEEE/CVF International Conference on
  Computer Vision}, pages 112--121, 2021.

\bibitem{cao2019learning}
Kaidi Cao, Colin Wei, Adrien Gaidon, Nikos Arechiga, and Tengyu Ma.
\newblock Learning imbalanced datasets with label-distribution-aware margin
  loss.
\newblock {\em Advances in neural information processing systems}, 32, 2019.

\bibitem{chawla2002smote}
Nitesh~V Chawla, Kevin~W Bowyer, Lawrence~O Hall, and W~Philip Kegelmeyer.
\newblock Smote: synthetic minority over-sampling technique.
\newblock {\em Journal of artificial intelligence research}, 16:321--357, 2002.

\bibitem{cheng2022compound}
Lechao Cheng, Chaowei Fang, Dingwen Zhang, Guanbin Li, and Gang Huang.
\newblock Compound batch normalization for long-tailed image classification.
\newblock In {\em Proceedings of the 30th ACM International Conference on
  Multimedia}, pages 1925--1934, 2022.

\bibitem{cui2021parametric}
Jiequan Cui, Zhisheng Zhong, Shu Liu, Bei Yu, and Jiaya Jia.
\newblock Parametric contrastive learning.
\newblock In {\em ICCV}, pages 715--724, 2021.

\bibitem{drummond2003c4}
Chris Drummond, Robert~C Holte, et~al.
\newblock C4. 5, class imbalance, and cost sensitivity: why under-sampling
  beats over-sampling.
\newblock In {\em Workshop on learning from imbalanced datasets II}, volume~11,
  pages 1--8. Citeseer, 2003.

\bibitem{he2016deep}
Kaiming He, Xiangyu Zhang, Shaoqing Ren, and Jian Sun.
\newblock Deep residual learning for image recognition.
\newblock In {\em Proceedings of the IEEE conference on computer vision and
  pattern recognition}, pages 770--778, 2016.

\bibitem{hong2021disentangling}
Youngkyu Hong, Seungju Han, Kwanghee Choi, Seokjun Seo, Beomsu Kim, and Buru
  Chang.
\newblock Disentangling label distribution for long-tailed visual recognition.
\newblock In {\em Proceedings of the IEEE/CVF conference on computer vision and
  pattern recognition}, pages 6626--6636, 2021.

\bibitem{huang2021scribble}
Peiliang Huang, Junwei Han, Nian Liu, Jun Ren, and Dingwen Zhang.
\newblock Scribble-supervised video object segmentation.
\newblock {\em IEEE/CAA Journal of Automatica Sinica}, 9(2):339--353, 2021.

\bibitem{kang2019decoupling}
Bingyi Kang, Saining Xie, Marcus Rohrbach, Zhicheng Yan, Albert Gordo, Jiashi
  Feng, and Yannis Kalantidis.
\newblock Decoupling representation and classifier for long-tailed recognition.
\newblock In {\em ICLR}, 2019.

\bibitem{krizhevsky2009learning}
Alex Krizhevsky, Geoffrey Hinton, et~al.
\newblock Learning multiple layers of features from tiny images.
\newblock 2009.

\bibitem{li2022trustworthy}
Bolian Li, Zongbo Han, Haining Li, Huazhu Fu, and Changqing Zhang.
\newblock Trustworthy long-tailed classification.
\newblock In {\em Proceedings of the IEEE/CVF Conference on Computer Vision and
  Pattern Recognition}, pages 6970--6979, 2022.

\bibitem{li2022long}
Mengke Li, Yiu-ming Cheung, and Yang Lu.
\newblock Long-tailed visual recognition via gaussian clouded logit adjustment.
\newblock In {\em Proceedings of the IEEE/CVF Conference on Computer Vision and
  Pattern Recognition}, pages 6929--6938, 2022.

\bibitem{li2022targeted}
Tianhong Li, Peng Cao, Yuan Yuan, Lijie Fan, Yuzhe Yang, Rogerio~S Feris, Piotr
  Indyk, and Dina Katabi.
\newblock Targeted supervised contrastive learning for long-tailed recognition.
\newblock In {\em CVPR}, pages 6918--6928, 2022.

\bibitem{lin2017focal}
Tsung-Yi Lin, Priya Goyal, Ross Girshick, Kaiming He, and Piotr Doll{\'a}r.
\newblock Focal loss for dense object detection.
\newblock In {\em Proceedings of the IEEE international conference on computer
  vision}, pages 2980--2988, 2017.

\bibitem{liu2019large}
Ziwei Liu, Zhongqi Miao, Xiaohang Zhan, Jiayun Wang, Boqing Gong, and Stella~X
  Yu.
\newblock Large-scale long-tailed recognition in an open world.
\newblock In {\em Proceedings of the IEEE/CVF Conference on Computer Vision and
  Pattern Recognition}, pages 2537--2546, 2019.

\bibitem{menon2020long}
Aditya~Krishna Menon, Sadeep Jayasumana, Ankit~Singh Rawat, Himanshu Jain,
  Andreas Veit, and Sanjiv Kumar.
\newblock Long-tail learning via logit adjustment.
\newblock In {\em ICLR}, 2021.

\bibitem{pan2022computer}
Chengwei Pan, Gangming Zhao, Junjie Fang, Baolian Qi, Jiaheng Liu, Chaowei
  Fang, Dingwen Zhang, Jinpeng Li, and Yizhou Yu.
\newblock Computer-aided tuberculosis diagnosis with attribute reasoning
  assistance.
\newblock In {\em International Conference on Medical Image Computing and
  Computer-Assisted Intervention}, pages 623--633. Springer, 2022.

\bibitem{pan2022learning}
Junwen Pan, Pengfei Zhu, Kaihua Zhang, Bing Cao, Yu Wang, Dingwen Zhang, Junwei
  Han, and Qinghua Hu.
\newblock Learning self-supervised low-rank network for single-stage weakly and
  semi-supervised semantic segmentation.
\newblock {\em International Journal of Computer Vision}, 130(5):1181--1195,
  2022.

\bibitem{park2021influence}
Seulki Park, Jongin Lim, Younghan Jeon, and Jin~Young Choi.
\newblock Influence-balanced loss for imbalanced visual classification.
\newblock In {\em Proceedings of the IEEE/CVF International Conference on
  Computer Vision}, pages 735--744, 2021.

\bibitem{ren2020balanced}
Jiawei Ren, Cunjun Yu, Xiao Ma, Haiyu Zhao, Shuai Yi, et~al.
\newblock Balanced meta-softmax for long-tailed visual recognition.
\newblock {\em Advances in neural information processing systems},
  33:4175--4186, 2020.

\bibitem{simonyan2014very}
Karen Simonyan and Andrew Zisserman.
\newblock Very deep convolutional networks for large-scale image recognition.
\newblock In {\em ICLR}, 2015.

\bibitem{tang2020long}
Kaihua Tang, Jianqiang Huang, and Hanwang Zhang.
\newblock Long-tailed classification by keeping the good and removing the bad
  momentum causal effect.
\newblock {\em Advances in Neural Information Processing Systems},
  33:1513--1524, 2020.

\bibitem{tang2022invariant}
Kaihua Tang, Mingyuan Tao, Jiaxin Qi, Zhenguang Liu, and Hanwang Zhang.
\newblock Invariant feature learning for generalized long-tailed
  classification.
\newblock In {\em ECCV}, 2022.

\bibitem{van2018inaturalist}
Grant Van~Horn, Oisin Mac~Aodha, Yang Song, Yin Cui, Chen Sun, Alex Shepard,
  Hartwig Adam, Pietro Perona, and Serge Belongie.
\newblock The inaturalist species classification and detection dataset.
\newblock In {\em Proceedings of the IEEE conference on computer vision and
  pattern recognition}, pages 8769--8778, 2018.

\bibitem{wang2022double}
Kuo Wang, Yuxiang Nie, Chaowei Fang, Chengzhi Han, Xuewen Wu, Xiaohui Wang,
  Liang Lin, Fan Zhou, and Guanbin Li.
\newblock Double-check soft teacher for semi-supervised object detection.
\newblock In {\em International Joint Conference on Artificial Intelligence
  (IJCAI)}, 2022.

\bibitem{wang2020long}
Xudong Wang, Long Lian, Zhongqi Miao, Ziwei Liu, and Stella~X Yu.
\newblock Long-tailed recognition by routing diverse distribution-aware
  experts.
\newblock In {\em ICLR}, 2021.

\bibitem{Xu2021TowardsCM}
Zhengzhuo Xu, Zenghao Chai, and Chun Yuan.
\newblock Towards calibrated model for long-tailed visual recognition from
  prior perspective.
\newblock In {\em NeurIPS}, pages 7139--7152, 2021.

\bibitem{yun2019cutmix}
Sangdoo Yun, Dongyoon Han, Seong~Joon Oh, Sanghyuk Chun, Junsuk Choe, and
  Youngjoon Yoo.
\newblock Cutmix: Regularization strategy to train strong classifiers with
  localizable features.
\newblock In {\em ICCV}, pages 6023--6032, 2019.

\bibitem{zhang2022generalized}
Dingwen Zhang, Guangyu Guo, Wenyuan Zeng, Lei Li, and Junwei Han.
\newblock Generalized weakly supervised object localization.
\newblock {\em IEEE Transactions on Neural Networks and Learning Systems},
  2022.

\bibitem{zhang2021weakly}
Dingwen Zhang, Junwei Han, Gong Cheng, and Ming-Hsuan Yang.
\newblock Weakly supervised object localization and detection: A survey.
\newblock {\em IEEE transactions on pattern analysis and machine intelligence},
  44(9):5866--5885, 2021.

\bibitem{zhang2018spftn}
Dingwen Zhang, Junwei Han, Le Yang, and Dong Xu.
\newblock Spftn: A joint learning framework for localizing and segmenting
  objects in weakly labeled videos.
\newblock {\em IEEE transactions on pattern analysis and machine intelligence},
  42(2):475--489, 2018.

\bibitem{zhang2019leveraging}
Dingwen Zhang, Junwei Han, Long Zhao, and Deyu Meng.
\newblock Leveraging prior-knowledge for weakly supervised object detection
  under a collaborative self-paced curriculum learning framework.
\newblock {\em International Journal of Computer Vision}, 127(4):363--380,
  2019.

\bibitem{zhang2020exploring}
Dingwen Zhang, Guohai Huang, Qiang Zhang, Jungong Han, Junwei Han, Yizhou Wang,
  and Yizhou Yu.
\newblock Exploring task structure for brain tumor segmentation from
  multi-modality mr images.
\newblock {\em IEEE Transactions on Image Processing}, 29:9032--9043, 2020.

\bibitem{zhang2021cross}
Dingwen Zhang, Guohai Huang, Qiang Zhang, Jungong Han, Junwei Han, and Yizhou
  Yu.
\newblock Cross-modality deep feature learning for brain tumor segmentation.
\newblock {\em Pattern Recognition}, 110:107562, 2021.

\bibitem{zhang2022onfocus}
Dingwen Zhang, Bo Wang, Gerong Wang, Qiang Zhang, Jiajia Zhang, Jungong Han,
  and Zheng You.
\newblock Onfocus detection: Identifying individual-camera eye contact from
  unconstrained images.
\newblock {\em Science China Information Sciences}, 65(6):1--12, 2022.

\bibitem{zhang2020weakly}
Dingwen Zhang, Wenyuan Zeng, Jieru Yao, and Junwei Han.
\newblock Weakly supervised object detection using proposal-and semantic-level
  relationships.
\newblock {\em IEEE Transactions on Pattern Analysis and Machine Intelligence},
  2020.

\bibitem{zhang2021automatic}
Dingwen Zhang, Jiajia Zhang, Qiang Zhang, Jungong Han, Shu Zhang, and Junwei
  Han.
\newblock Automatic pancreas segmentation based on lightweight dcnn modules and
  spatial prior propagation.
\newblock {\em Pattern Recognition}, 114:107762, 2021.

\bibitem{zhang2017mixup}
Hongyi Zhang, Moustapha Cisse, Yann~N Dauphin, and David Lopez-Paz.
\newblock mixup: Beyond empirical risk minimization.
\newblock In {\em ICLR}, 2018.

\bibitem{zhang2021test}
Yifan Zhang, Bryan Hooi, Lanqing Hong, and Jiashi Feng.
\newblock Test-agnostic long-tailed recognition by test-time aggregating
  diverse experts with self-supervision.
\newblock In {\em NeurIPS}, 2022.

\bibitem{zhao2021contralaterally}
Gangming Zhao, Chaowei Fang, Guanbin Li, Licheng Jiao, and Yizhou Yu.
\newblock Contralaterally enhanced networks for thoracic disease detection.
\newblock {\em IEEE Transactions on Medical Imaging}, 40(9):2428--2438, 2021.

\bibitem{zhao2021weakly}
Wangbo Zhao, Jing Zhang, Long Li, Nick Barnes, Nian Liu, and Junwei Han.
\newblock Weakly supervised video salient object detection.
\newblock In {\em Proceedings of the IEEE/CVF conference on computer vision and
  pattern recognition}, pages 16826--16835, 2021.

\bibitem{zhao2022cross}
Xinkai Zhao, Chaowei Fang, De-Jun Fan, Xutao Lin, Feng Gao, and Guanbin Li.
\newblock Cross-level contrastive learning and consistency constraint for
  semi-supervised medical image segmentation.
\newblock In {\em 2022 IEEE 19th International Symposium on Biomedical Imaging
  (ISBI)}, pages 1--5. IEEE, 2022.

\bibitem{zhao2021deep}
Xinkai Zhao, Chaowei Fang, Feng Gao, FAN De-Jun, Xutao Lin, and Guanbin Li.
\newblock Deep transformers for fast small intestine grounding in capsule
  endoscope video.
\newblock In {\em 2021 IEEE 18th International Symposium on Biomedical Imaging
  (ISBI)}, pages 150--154. IEEE, 2021.

\bibitem{zhao2022adaptive}
Yan Zhao, Weicong Chen, Xu Tan, Kai Huang, and Jihong Zhu.
\newblock Adaptive logit adjustment loss for long-tailed visual recognition.
\newblock In {\em AAAI}, volume~36, pages 3472--3480, 2022.

\bibitem{zhou2020bbn}
Boyan Zhou, Quan Cui, Xiu-Shen Wei, and Zhao-Min Chen.
\newblock Bbn: Bilateral-branch network with cumulative learning for
  long-tailed visual recognition.
\newblock In {\em Proceedings of the IEEE/CVF conference on computer vision and
  pattern recognition}, pages 9719--9728, 2020.

\end{thebibliography}

}

\end{document}